\documentclass[letterpaper]{article} 
\usepackage{aaai2026}  
\usepackage{times}  
\usepackage{helvet}  
\usepackage{courier}  
\usepackage[hyphens]{url}  
\usepackage{graphicx} 
\usepackage{natbib}  
\usepackage{caption} 
\usepackage{algorithm}

\usepackage[noend]{algpseudocode}
\usepackage{amsmath,amsfonts,amsthm}
\usepackage{adjustbox}
\usepackage{textcomp}
\usepackage{subcaption}
\usepackage{tabularx}
\usepackage{booktabs}
\usepackage{multirow}
\usepackage{makecell}
\usepackage{amssymb}

\newtheorem{manualthminner}{Theorem}
\newenvironment{manualthm}[2][]{
  \renewcommand\themanualthminner{#2}
  \manualthminner[#1]
}{%
  \endmanualthminner
}

\newtheorem{manualleminner}{Lemma}
\newenvironment{manuallem}[2][]{
  \renewcommand\themanualleminner{#2}
  \manualleminner[#1]
}{%
  \endmanualleminner
}

\newtheorem{manualtheorem}{Theorem}

\newtheorem{manuallemma}{Lemma}

\usepackage{newfloat}
\usepackage{listings}
\DeclareCaptionStyle{ruled}{labelfont=normalfont,labelsep=colon,strut=off} 
\floatstyle{ruled}
\newfloat{listing}{tb}{lst}{}
\floatname{listing}{Listing}
\title{Sheaf Graph Neural Networks via PAC-Bayes Spectral Optimization}

\author{
    Yoonhyuk Choi$^1$, Jiho Choi$^2$, Taewook Ko$^3$ \\ JongWook Kim$^4$, Chong-Kwon Kim$^5$
}
\affiliations{
    \textsuperscript{1}Sookmyung Women's University, Seoul, Republic of Korea\\

    \textsuperscript{2}Korea Advanced Institute of Science and Technology (KAIST), Seoul, Republic of Korea\\
    
    \textsuperscript{3}Samsung Electronics, Seoul, Republic of Korea\\

    \textsuperscript{4}Sangmyung University, Seoul, Republic of Korea\\
    
    \textsuperscript{5}Korea Institute of Energy Technology (KENTECH), Naju, Republic of Korea\\
    chldbsgur123@sookmyung.ac.kr, 
    jihochoi1993@gmail.com,
    taewook.ko@snu.ac.kr,
    jkim@smu.ac.kr,
    ckim@kentech.ac.kr
}

\begin{document}

\maketitle

\begin{abstract}
Over-smoothing in Graph Neural Networks (GNNs) causes collapse in distinct node features, particularly on heterophilic graphs where adjacent nodes often have dissimilar labels. Although sheaf neural networks partially mitigate this problem, they typically rely on static or heavily parameterized sheaf structures that hinder generalization and scalability. Existing sheaf-based models either predefine restriction maps or introduce excessive complexity, yet fail to provide rigorous stability guarantees. In this paper, we introduce a novel scheme called SGPC (Sheaf GNNs with PAC-Bayes Calibration), a unified architecture that combines cellular-sheaf message passing with several mechanisms, including optimal transport-based lifting, variance-reduced diffusion, and PAC-Bayes spectral regularization for robust semi-supervised node classification. We establish performance bounds theoretically and demonstrate that end-to-end training in linear computational complexity can achieve the resulting bound-aware objective. Experiments on nine homophilic and heterophilic benchmarks show that SGPC outperforms state-of-the-art spectral and sheaf-based GNNs while providing certified confidence intervals on unseen nodes. The code and proofs are in https://github.com/ChoiYoonHyuk/SGPC.
\end{abstract}

\section{Introduction}\label{sec:intro}
The explosive growth of graph-structured data across social \cite{fan2019graph}, biological \cite{zhang2021graph}, and industrial domains \cite{chen2021interaction} has established Graph Neural Networks (GNNs) as a cornerstone of modern machine learning. Classic message passing GNNs  \cite{kipf2016semi,velickovic2017graph,defferrard2016convolutional} aggregate neighbor signals under an implicit homophily assumption, where adjacent nodes tend to share labels or attributes. The resulting low-pass filters perform Laplacian smoothing \cite{li2022finding}, which is effective on homophilic graphs but provably degrades under heterophily or adversarial structure \cite{pei2020geom,zhu2020beyond}. While recent spatial remedies like edge re-weighting \cite{choi2025beyond}, subgraph sampling \cite{bo2021beyond}, and attention mechanisms \cite{brody2021attentive} yield empirical gains, they treat edges as scalar weights and largely ignore uncertainty.

\begin{figure}[t]
\centering
 \includegraphics[width=.45\textwidth]{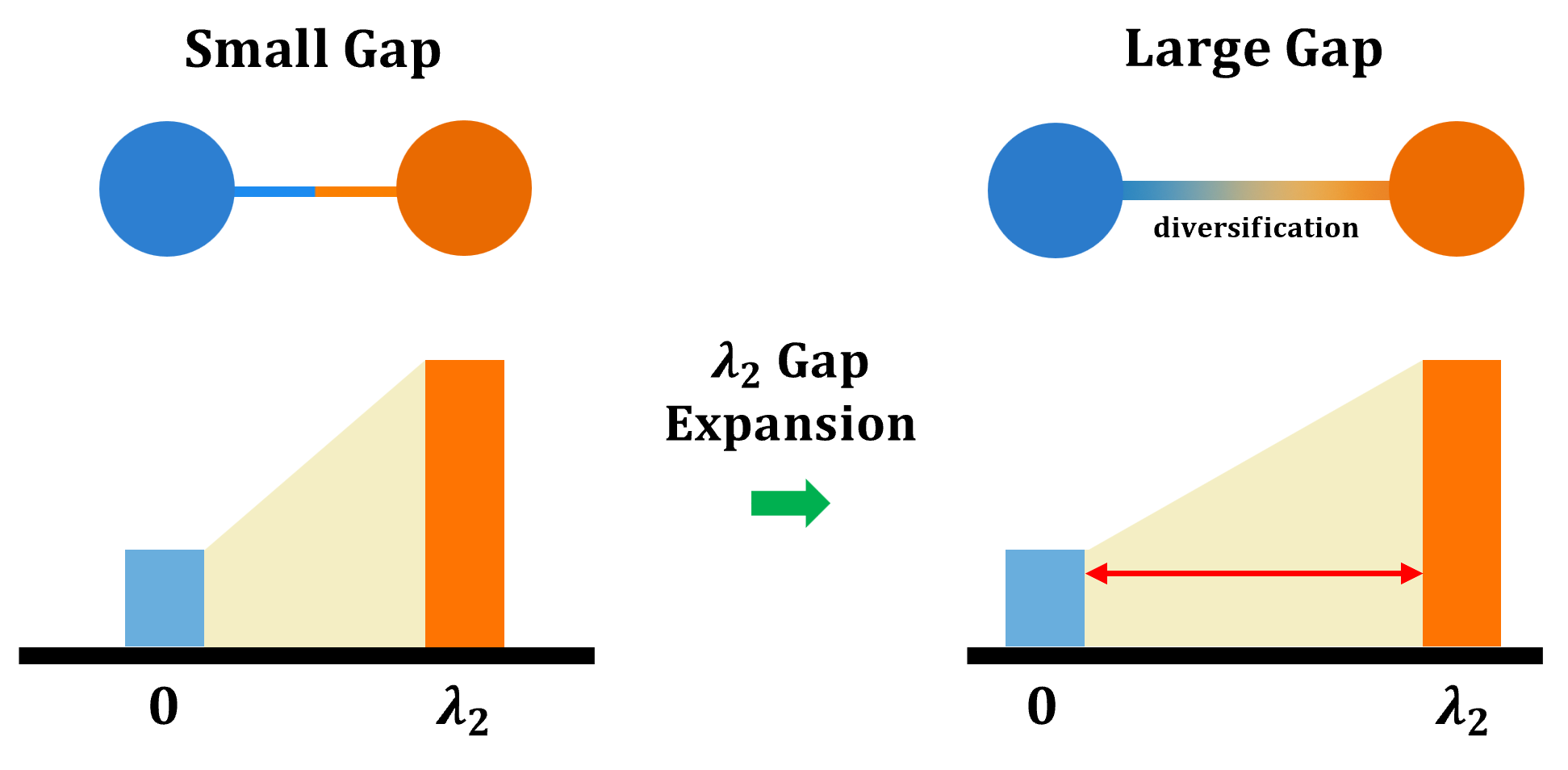}
    \caption{In the small-gap regime (left), two nodes are connected by a weak edge, so the Laplacian spectrum shows only a narrow separation between the first eigenvalue (0) and the second one ($\lambda_2$). After $\lambda_2$-gap expansion (right), the edge becomes strong and smoothly color-graded with a wide spectrum, illustrating the enlarged spectral gap}
  \label{fig_intro}
\end{figure}

Cellular sheaf theory reinterprets an edge as a linear restriction between local feature spaces, inducing a sheaf Laplacian whose spectrum captures edge directionality and class dispersion \cite{hansen2020sheaf, bodnar2022neural}. Previous studies showed that matrix edge representation can suppress over-smoothing while respecting feature anisotropy. However, existing sheaf GNNs suffer from several limitations: they (i) fix restriction maps via simple gating mechanisms, (ii) lack heterophily-aware uncertainty calibration, and (iii) offer no generalization guarantees beyond empirical test accuracy \cite{zaghen2024sheaf}. These gaps hinder widespread adoption across various domains, highlighting the need for calibrated risk and theoretical robustness.

As one solution, PAC-Bayes analysis offers a principled route by linking expected risk to posterior-prior compression and data-dependent margins \cite{zhou2018non,letarte2019dichotomize}. Yet current PAC-Bayes bounds for GNNs remain loose because they ignore the spectrum of the underlying operator (e.g., the sheaf or graph Laplacian), which governs diffusion depth and representational expressivity \cite{xu2018powerful}. A tighter, spectrum-aware bound can directly translate spectral engineering into certified risk reduction.

Motivated by these gaps, we propose Sheaf Graph Neural Networks with PAC-Bayes Calibration (SGPC), a unified framework that learns sheaf restrictions, calibrates their uncertainty, and optimizes the spectral gap under a provably tight PAC-Bayes bound. As illustrated in Figure \ref{fig_intro}, spectral gap optimization diversifies sheaves and separates informative frequency components from near‑null modes. This can enhance the model’s ability to discriminate labels across classes and further tighten the generalization bound. Our contribution can be summarized as follows:
\begin{itemize}
  \item We propose SGPC, a fully differentiable architecture that learns sheaf restriction maps via a Wasserstein-Entropic Lift, which optimizes the sheaf Laplacian spectral gap.
  \item We derive theoretical guarantees for cellular-sheaf GNNs, including convergence, spectral gap increase with risk control, and generalization bound.
  \item Extensive experiments on nine benchmarks demonstrate that SGPC outperforms state-of-the-art GNNs and prior sheaf models, while providing PAC-Bayes calibrated uncertainty estimates and provably tighter risk bounds.
\end{itemize}

\section{Related Work}\label{sec:related}

\paragraph{Heterophilic Graph Neural Networks.} Spectral formulations such as GCN \cite{kipf2016semi}, ChebNet \cite{defferrard2016convolutional}, and spatial attention models like GAT \cite{velickovic2017graph} rely on low-pass filters, which perform well under high homophily. Subsequent works improved depth and scalability but largely retained the homophily prior. Early solutions attempted to decouple ego and neighbor features \cite{zhu2020beyond} or to sample distant yet similar nodes \cite{pei2020geom}. Recent surveys catalog more than 50 heterophily-oriented architectures \cite{luan2024heterophilic}. Notable trends include edge reweighting \cite{choi2025beyond}, causal discovery for message routing \cite{wang2025heterophilic}, and adaptive frequency mixing \cite{choi2025specsphere}. Despite this progress, many methods overlook signed structures or treat all disassortative edges uniformly, underscoring the need for a sheaf-aware neural network capable of handling heterophily.

\paragraph{Sheaf Theory and Spectral Optimization.} Neural Sheaf Diffusion (NSD) \cite{hansen2020sheaf} advances graph representation learning by endowing graphs with non‑trivial cellular sheaves. On the empirical side, sheaf Laplacian‑based GNN models consistently outperform baseline GCNs on signed and heterophilic benchmarks, demonstrating clear gains in classification accuracy \cite{barbero2022sheaf}. More recent efforts have extended NSD to handle nonlinear sheaf Laplacians that capture complex interactions \cite{zaghen2024nonlinear}. Recent works have begun complementing sheaf-based diffusion with spectral optimization techniques \cite{hansen2019toward}. For example, \cite{bodnar2022neural} demonstrates that the spectral gap is tightly linked to path-dependent transport maps, optimizing this via path alignment. Others incorporate directional bias through a directed cellular sheaf, deriving a directed Laplacian to improve task-specific bias \cite{duta2023sheaf}.

\paragraph{Summary \& Gap.} Although existing methods effectively model heterophilic structures, they often face limitations in scalability, spectral control, or generality. In contrast, we (i) employ a joint diffusion model \cite{caralt2024joint} that learns restriction maps and features concurrently, reducing parameter count while preserving inductive bias; (ii) incorporate a spectral gap regularization term during training, ensuring better control over diffusion stability and linear separability; (iii) learn asymmetric sheaf maps under spectral constraints, enabling task-adaptive directionality while bounding eigenvalue distributions.

\section{Preliminaries}\label{sec:prelim}
Let $\mathcal{G} = (\mathcal{V}, \mathcal{E}, H)$ denote an attribute graph with $n = |\mathcal{V}|$ nodes and $m = |\mathcal{E}|$ edges. The node feature matrix $H \in \mathbb{R}^{n \times d_0}$ encodes $d_0$-dimensional input vectors for each node. The adjacency matrix $A \in \{0, 1\}^{n \times n}$ captures the edge structure of $\mathcal{G}$, and $D$ is a diagonal degree matrix with entries $d_{ii} = \sum_{j=1}^n A_{ij}$. Each node is associated with a one-hot label vector in $Y \in \mathbb{R}^{n \times C}$, where $C$ is the number of classes. To quantify class consistency along edges, we define the global edge homophily ratio $\mathcal{G}_h$ as follows:
\begin{equation}
\label{global_homo}
\mathcal{G}_h := \frac{1}{|\mathcal{E}|} \sum_{(i,j) \in \mathcal{E}} \mathbb{I}(Y_i = Y_j),
\end{equation}
where $\mathbb{I}(\cdot)$ is the indicator function. Given a labeled subset $\mathcal{V}_L \subset \mathcal{V}$, the task of semi-supervised node classification is to infer the class labels of the remaining unlabeled nodes $\mathcal{V}_U := \mathcal{V} \setminus \mathcal{V}_L$. The following section introduces the cellular sheaf and geometric foundations.

\paragraph{Graphs and Cellular Sheaves.}
In graph $\mathcal{G}$, each vertex $i$ carries an input feature $h_i\in\mathbb R^{d_0}$ and a label $y_i\in\{1,\dots,C\}$. Throughout the paper, we fix $d_0$ as the input feature dimension and $d$ as the sheaf-fibre dimension. A cellular sheaf $\mathcal F$ over $\mathcal{G}$ assigns a fibre $\mathcal F(v_i)=\mathbb R^{d}$ to every vertex and $\mathcal F(e_{ij})=\mathbb R^{d}$ to every edge, together with linear restriction maps $R_{ij}:\mathcal F(v_i)\to\mathcal F(e_{ij})$ and $R_{ji}$ in the opposite direction. The sheaf incidence matrix is the block matrix $B\in\mathbb R^{md\times nd}$ whose $(e_{ij},v_i)$ block is $R_{ij}$ and $-R_{ji}$ for $(e_{ij},v_j)$.

\paragraph{Optimal Transport and Wasserstein Geometry.}
Node features are interpreted as empirical probability measures in the 2-Wasserstein space $(\mathcal P_2(\mathbb R^{d_0}), W_2)$. Here, $\mathcal P_2(\mathbb R^{d_0})$ denotes the set of probability measures on $\mathbb R^{d_0}$ with finite second moments, and $W_2$ is the associated 2-Wasserstein distance. Given node features $h_i,h_j\in\mathbb R^{d_0}$, we define $\mu=h_i/\|h_i\|_1$ and $\nu=h_j/\|h_j\|_1$ as empirical measures in $\mathcal P_2(\mathbb R^{d_0})$ by $\ell_1$-normalization. Let $e_p\in\mathbb R^{d_0}$ denote the $p$-th canonical basis vector, i.e.\ $(e_p)_\ell = \mathbf 1\{\ell=p\}$. Then, the canonical-basis cost is given by:
\begin{equation}\label{eq_cfeat}
   C_{\mathrm{feat}}[p,q]=\|e_p-e_q\|_2^2.
\end{equation}
Consequently, the entropic optimal transport (OT) problem can be defined as follows:
\begin{equation}\label{eq_sinkhorn}
P_\star=\arg\min_{P\in\Pi(\mu,\nu)}
\langle P,C_{\mathrm{feat}}\rangle + \varepsilon\mathcal H(P),
\end{equation}
where $\mathcal H(P)$ denotes the Shannon entropy $\mathcal H(P) = -\sum_{ij} P_{ij}\log P_{ij}$, encouraging smoother (high-entropy) couplings.

\begin{figure*}[t]
\centering
 \includegraphics[width=\textwidth]{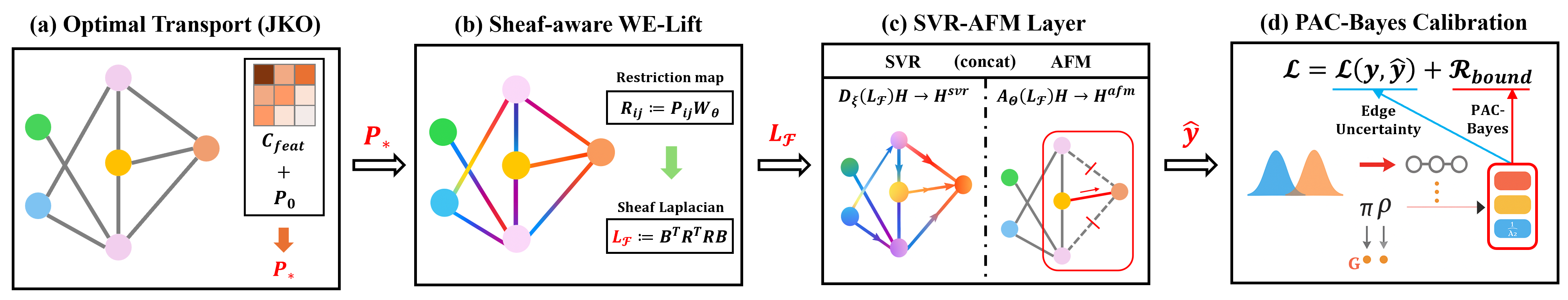}
    \caption{(a) A Jordan-Kinderlehrer-Otto (JKO) step refines the initial Sinkhorn plan $P_{0}$ under the feature-cost matrix $C_{\mathrm{feat}}$, producing a globally stable coupling $P_{*}$; (b) The coupling $P_{ *}$ is turned into restriction maps $R_{ij}$, which in turn define the sheaf Laplacian $L_{\mathcal F}$; (c) Stochastic variance-reduced diffusion $D_{\xi}$ with adaptive frequency mixing $A_\theta$ yields node-level predictions $\hat y$; (d) A $\beta$-Dirichlet posterior calibrates edge uncertainty, while an optimizer enlarges the spectral gap $\lambda_{2}$}
  \label{model}
\end{figure*}

\section{Methodology}\label{sec:method}
We introduce Sheaf GNNs with PAC-Bayes Calibration (SGPC), which learns graph-structured sheaf parameters while providing a PAC-Bayes generalization bound for cellular-sheaf GNNs as follows:
\begin{enumerate}
    \item \textbf{OT with Sheaf-aware WE Lift} refines Sinkhorn optimal transport via a single-step Wasserstein gradient flow, producing globally stable restriction maps.
    \item \textbf{SVR-AFM Layer} combines (i) Stochastic Variance-Reduced (SVR) diffusion for global denoising and (ii) an Adaptive Frequency Mixing (AFM) branch.
    \item \textbf{PAC-Bayes Spectral Optimization} jointly (i) calibrates edge-level uncertainty through a $\beta$-Dirichlet model with explicit heterophily penalties and (ii) tightens the PAC-Bayes bound by optimizing the sheaf Laplacian spectral gap~$\lambda_2$ under a perturbation constraint.
\end{enumerate}

\subsection{OT with Sheaf-aware WE Lift}\label{sec:wasserlift}
As shown in Figure \ref{model}-(a), we generalize the classic Sinkhorn coupling (Eq. \ref{eq_sinkhorn}) by evolving the transport plan $P_t$ (step $t$) inside the $2$-Wasserstein metric space:
\begin{equation}
  \partial_t P_t = -\nabla_{W_2} \Bigl[\langle P_t,C_{\mathrm{feat}}\rangle
  +\varepsilon\mathcal H(P_t)\Bigr],
\end{equation}
where $C_{\mathrm{feat}}$ (Eq. \ref{eq_cfeat}) is the pairwise feature cost and $\mathcal H(\cdot)$ is Shannon entropy. Starting from $P_0=P_{\text{Sinkhorn}}$, a Jordan-Kinderlehrer-Otto (JKO) step refines the transport plan towards a globally KL‑stable configuration as below:
\begin{equation}
\label{eq_varepsilon}
    P_\star = \arg\min_{P}\left\{\frac{1}{2} W_2^2(P, P_0) + \langle P, C_{\mathrm{feat}} \rangle + \varepsilon \mathcal{H}(P)\right\},
\end{equation}
As in Figure \ref{model}-(b), the restriction maps $R_{ij}$ are generated using the refined transport plan $P_*$ as follows:
\begin{equation}\label{eq_r}
  R_{ij} := P_{\star,ij} W_\theta,\quad W_\theta \in \mathbb{R}^{d_0 \times d_0}.
\end{equation}
Although $P_\star$ is obtained via OT, it is still updated end-to-end by the node-classification loss, making $R_{ij}$ task-adaptive and differentiable. These maps populate the block-diagonal tensor and enter the sheaf Laplacian $L_{\mathcal F}$ in Eq. \ref{eq:Lf_pre}.

\subsection{SVR-AFM Layer}
The collection $\{R_{ij}\}$ endows the graph with a cellular-sheaf structure whose co-boundary matrix is $B\in\mathbb R^{|E|\times|V|}$ (edge-to-node relationship). Let $R := \operatorname{diag}(R_{ij})$ and define the sheaf Laplacian as below:
\begin{equation}\label{eq:Lf_pre}
  L_{\mathcal F}:=(B\otimes I_{d_0})^\top R^\top R (B\otimes I_{d_0})
\end{equation}
When each $R_{ij}$ collapses to a scalar weight, Eq. \ref{eq:Lf_pre} reduces to the standard graph Laplacian. Given node features $H \in \mathbb R^{n\times d_0}$, diffusion hyperparameters $\xi=(\Delta t)$, and frequency mixing weights $\Theta$, we design the SVR-AFM pipeline below.

\paragraph{Stochastic variance-reduced (SVR) diffusion.}
As shown in the left side of Figure \ref{model}-(c), we introduce the diffusion process using the sheaf Laplacian $L_{\mathcal F}$ (Eq. \ref{eq:Lf_pre}) below:
\begin{equation}\label{eq:svrg_diff}
  H^{\mathrm{svr}}
  =\mathcal D_{\xi}(L_{\mathcal F})H
  \approx (I+\Delta t L_{\mathcal F})^{-1}H,
\end{equation}
where a few SVR-preconditioned conjugate-gradient iterations approximate the inverse. Further details on solving this iteration are provided in Eq. \ref{eq_cg_solve}.

\paragraph{Adaptive Frequency Mixing (AFM) branch.}
Let $Q$ denote the maximum polynomial order and $T_q(\cdot)$ the $q$-th Chebyshev polynomial. As in the right side of the Figure \ref{model}-(c), we utilize learnable frequency coefficients below:
\begin{equation}
\alpha_q=\frac{\exp(\gamma_q)}{\sum_{p=0}^{Q}\exp(\gamma_p)}, 
\quad q=0,\ldots,Q,
\end{equation}
where $\gamma_q\in\mathbb{R}$ are free parameters. The AFM representation is then obtained as follows:
\begin{equation}
H^{\mathrm{afm}} =  \mathcal A_\Theta\bigl(L_{\mathcal F}\bigr)H = \sum_{q=0}^{Q}\alpha_q\,T_q \bigl(\widetilde L\bigr)H,
\end{equation}
with $\widetilde L = I - D^{-1/2} L_{\mathcal F} D^{-1/2}$ the symmetrically normalized sheaf Laplacian. The term $q=0$ corresponds to the identity operator (no filtering). Higher-order terms $q\ge1$ serve as polynomial bases that can approximate both low- and high-pass behaviors depending on the learned coefficients $\{\alpha_q\}$. Consequently, on heterophilous graphs, the model tends to place more weight on combinations whose spectral response is larger at high eigenvalues (i.e., high-frequency components).

\paragraph{Branch fusion.}
We first concatenate the outputs from the sheaf-based diffusion branch and the AFM branch, and feed the result to a lightweight projector with two layers. Then we apply an MLP or GAT \cite{velickovic2017graph} to mix channels and propagate along the graph:
\begin{equation}
\label{eq_fusion}
H' = F_{\text{mix}}([H^{\mathrm{svr}} || H^{\mathrm{afm}}]).
\end{equation}
Regarding $F_{\text{mix}}$, we employ a GAT module for homophilic datasets and an MLP for heterophilic ones. Given the fused representations $H'$ in Eq. \ref{eq_fusion}, the class probability of each node can be inferred as follows:
\begin{equation}
\label{eq_pred}
\widehat y = \operatorname{softmax} \bigl(W H'\bigr).
\end{equation}

\noindent
\textbf{Computational cost} \textit{is introduced in Appendix A.}

\subsection{PAC-Bayes Calibration} \label{sec:pacbayes}
The stochastic restriction maps $\{R_{ij}\}$ yielded by the WE Lift render every edge uncertain. We convert this uncertainty into a data-dependent posterior and then actively enlarge the sheaf spectral gap, obtaining a PAC-Bayes bound.

\paragraph{$\beta$-Dirichlet prior.}
For each edge, we model the message agreement rate $\kappa_{ij} \in [0,1]$ with the following prior:
\begin{equation}
\label{eq_pri}
\kappa_{ij} \sim \operatorname{Beta}(\alpha_{ij},\beta_{ij}),\quad 
\pi=\prod_{(i,j)\in\mathcal E}  \operatorname{Beta}(\alpha_{ij},\beta_{ij}).
\end{equation}

\paragraph{Posterior update.}
A three-step fixed-point solver updates the posterior for each edge $(i,j)$. Starting from the prior, the solver iteratively (i) computes pseudo-counts of agreement based on SVR-AFM output, (ii) calibrates them through a class-coupling matrix, and (iii) normalizes the posterior to avoid over-confidence. This process starts with fixed-point updates as follows:
\begin{equation}
\label{eq_pos}
\bar\kappa_{ij}:=\mathbb E_{\rho}[\kappa_{ij}]
=\frac{\bar\alpha_{ij}}{\bar\alpha_{ij}+\bar\beta_{ij}},\quad
\rho=\prod_{(i,j)\in\mathcal{E}}\operatorname{Beta}(\bar\alpha_{ij},\bar\beta_{ij}).
\end{equation}

\paragraph{Empirical risk optimization.}
The retrieved posterior mean $\bar \kappa_{ij} :=\mathbb E_{\rho}[\kappa_{ij}]$ serves as an edge uncertainty weight in calibrating empirical risk as below:
\begin{equation}
    f(\widehat{y}_i; \bar \kappa_{ij}) = \bar \kappa_{ij} \cdot \widehat{y}_i + (1-\bar \kappa_{ij}) \cdot y^{\mathrm{prior}},
\end{equation}
where the $y^{\mathrm{prior}}$ is a pre-defined class distribution (e.g., uniform). Given the class probability $\widehat y$ (Eq. \ref{eq_pred}), the calibrated empirical loss is given by:
\begin{equation}\label{eq_emp_risk}
\boxed{\mathcal {L}(y, \widehat y)
=\frac{1}{|\mathcal V_L|} \sum_{i\in\mathcal V_L} C\bigl(y_i, f \bigl(\widehat{y}_i; \bar \kappa_{ij}\bigr)\bigr)}
\end{equation}

\paragraph{KL-divergence.} To regularize the posterior complexity and control generalization, we define the divergence between the prior $\pi$ (Eq. \ref{eq_pri}) and posterior $\rho$ (Eq. \ref{eq_pos}):
\begin{equation}
\label{eq_kldiv}
    \boxed{\mathcal L_{\mathrm{KL}}=
\sqrt{\frac{\mathrm{KL}(\rho\Vert\pi)+\log(2/\delta)}{2n}}}
\end{equation}
where $n:=|\mathcal V_L|$ is the number of samples and the $\log(2/\delta)$ is confidence adjustment in PAC-Bayes bound.

\paragraph{Spectral-gap optimization.}
To isolate the heterophily effect that will appear in the bound, let $\mathcal E_{cc'}=\{(i,j):y_i=c, y_j=c'\}$. The posterior class-coupling matrix and its Frobenius norm are given by:
\begin{equation}
\label{eq_chet}
\Pi_{cc'}=\frac{1}{|\mathcal E_{cc'}|}
          \sum_{(i,j)\in\mathcal E_{cc'}}\bar \kappa_{ij},
\quad
c_{\mathrm{het}}=\|\Pi\|_{F}.
\end{equation}

\begin{algorithm}[t]\label{alg:SGPC}
  \caption{\textsc{SGPC}: One Training Epoch}
  \begin{algorithmic}[1]
    \Require Graph $\mathcal{G}=(\mathcal{V},\mathcal{E})$, node features $H$, labels $Y$, parameters $\theta$, hyperparameters $(\lambda_{\mathrm{KL}}, \lambda_{\mathrm{spec}})$
    \Ensure Updated $\theta$

    \Statex \textbf{(1) OT $\rightarrow$ WE Lift $\rightarrow$ Sheaf Laplacian}
      \State $P_0 \leftarrow \textsc{Sinkhorn}(H,\varepsilon)$
      \State $P_* \leftarrow \textsc{JKO}(P_0, C_{\mathrm{feat}}, \varepsilon)$
      \State $R_{ij} \leftarrow P_{*,ij} W_\theta$
      \State $L_{\mathcal F} \leftarrow (B\otimes I_{d_0})^\top \operatorname{diag}(R_{ij})^{\top} \operatorname{diag}(R_{ij})(B\otimes I_{d_0})$

    \Statex \textbf{(2) SVR-AFM Forward Pass}
      \State $H^{\mathrm{svr}} \leftarrow \textsc{SVR}(H, L_{\mathcal F})$
      \State $H^{\mathrm{afm}} \leftarrow \textsc{AFM}(H, L_{\mathcal F})$
      \State $H' \leftarrow F_{\text{mix}}([H^{\mathrm{svr}} \| H^{\mathrm{afm}}])$
      \State $\widehat{y} \leftarrow \operatorname{softmax}(W H')$

    \Statex \textbf{(3) $\boldsymbol\beta$-Dirichlet Posterior Update}
      \ForAll{$(i,j) \in \mathcal{E}$ \textbf{parallel}}
         \State $(\bar\alpha_{ij}, \bar\beta_{ij}) \leftarrow \textsc{FixedPoint}(\alpha_{ij}, \beta_{ij})$
         \State $\bar\kappa_{ij} \leftarrow \bar\alpha_{ij}/(\bar\alpha_{ij} + \bar\beta_{ij})$
      \EndFor

    \Statex \textbf{(4) PAC-Bayes Calibration \& Parameter Update}
      \State $\mathcal L(y,\widehat y) \leftarrow \frac{1}{|\mathcal V_{L}|} \sum_{i\in \mathcal V_{L}} C\bigl(y_i, f(\widehat{y}_i; \bar\kappa_{i\cdot})\bigr)$
      \State $\mathcal L_{\mathrm{KL}} \leftarrow \sqrt{\frac{\mathrm{KL}(\rho\Vert\pi)+\log(2/\delta)}{2|\mathcal V_{L}|}}$
      \State $\mathcal L_{\mathrm{spec}} \leftarrow c_{\mathrm{het}}/\lambda_2(L_{\mathcal F})$
      \State $\mathcal R_{\text{bound}} \leftarrow \lambda_{\mathrm{KL}} \mathcal L_{\mathrm{KL}} + \lambda_{\mathrm{spec}} \mathcal L_{\mathrm{spec}}$
      \State $\mathcal L \leftarrow \mathcal L(y,\widehat y) + \mathcal R_{\text{bound}}$
      \State $g \leftarrow \nabla_{L_{\mathcal F}} \lambda_2(L_{\mathcal F})$
      \State $L_{\mathcal F} \leftarrow L_{\mathcal F} + \eta g$
      \State $\{R_{ij}\} \leftarrow \textsc{ReassembleFrom}(L_{\mathcal F}, B)$
  \end{algorithmic}
\end{algorithm}

Let $\lambda_2$ stand for the spectral gap. For a stable diffusion, we introduce to enlarge $\lambda_2$ as below:
\begin{equation}
\label{eq_lambda}
    \lambda_2(L_{\mathcal{F}}) = \min_{v \perp \mathbf{1}} \frac{v^\top L_{\mathcal{F}} v}{v^\top v}.
\end{equation}
Combining the heterophily penalty $c_{\mathrm{het}}$ (Eq. \ref{eq_chet}) and the spectral gap $\lambda_2$ (Eq. \ref{eq_lambda}) jointly characterizes the stability of sheaf diffusion in heterophilic regimes as below: 
\begin{equation}
\label{eq_rspec}
    \boxed{\mathcal{L}_{\mathrm{spec}} = \frac{c_{\mathrm{het}}}{\lambda_2(L_{\mathcal{F}})}}
\end{equation}
which penalizes excessive heterophily relative to the diffusion capacity of the sheaf Laplacian.

\paragraph{Overall loss function.}
Given the calibrated cross-entropy $\mathcal{L}(y,\widehat y)$ (Eq. \ref{eq_emp_risk}), KL divergence $\mathcal{L}_{\mathrm{KL}}$ (Eq. \ref{eq_kldiv}), and spectral gap $\mathcal{L}_{\mathrm{spec}}$ (Eq. \ref{eq_rspec}), we can define the total loss as below:
\begin{equation}
\label{eq_loss}
    \mathcal L = \mathcal {L}(y, \widehat y) + \underbrace{
    \lambda_{\mathrm{KL}} \mathcal{L}_{\mathrm{KL}} + \lambda_{\mathrm{spec}} \mathcal{L}_{\mathrm{spec}}}_{\displaystyle\mathcal R_{\text{bound}}}
\end{equation}

\begin{manualtheorem}[PAC-Bayes Sheaf Generalization Bound]\label{thm:bound}
For any $\delta>0$, our model $f$ meets the following inequality over the data distribution $\mathcal D$ with probability at least $1-\delta$:
\begin{equation}\label{eq_lambda2_new}
\mathcal L_{\mathcal D}(f)
 \le \mathcal {L}(y, \widehat y) + \mathcal R_{\text{bound}}.
\end{equation}
\paragraph{Proof.} Please see Appendix B.
\end{manualtheorem}

\section{Theoretical Analysis}\label{sec:theory}
We analyze SGPC along four axes: (i) solver convergence, (ii) spectral-gap monotonicity, (iii) risk-variance contraction, and (iv) PAC-Bayes generalization. Throughout, we set $L_t := B^{\top}R_t^{\top}R_tB$ (the sheaf Laplacian at epoch $t$) and write $\lambda_{\max,t} := \lambda_{\max}(L_t)$ and $\lambda_{2,t} := \lambda_2(L_t)$.

\subsection{Convergence Analysis}\label{sec:conv}
To solve Eq. \ref{eq:svrg_diff}, we need to approximate the linear system at every SGPC layer as below:
\begin{equation}
\label{eq_cg_solve}
  \bigl(I+\Delta t L_t\bigr)h=b,
\end{equation}
where $b$ denotes the input node features from the previous layer. Running Conjugate Gradient (CG) directly on the dense sheaf Laplacian $L_t$ would cost $O \bigl(|E|\sqrt{\kappa(L_t)}\bigr)$ per solve $\bigl(\kappa(L_t) := \lambda_{\max}(L_t) / \lambda_{\min}(L_t)\bigr)$ and gives no iteration bound that is uniform throughout training. To guarantee epoch-wise scalability, we replace $L_t$ by a leverage-score spectral sparsifier $\tilde L_t$ containing only $O \bigl(|V|\log|V|/\varepsilon^{2}\bigr)$ non-zeros, and show that CG on the shifted system $\bigl(I+\Delta t \tilde L_t\bigr)h=b$ converges in $O\bigl(\log(1/\epsilon_{\mathrm{CG}})\bigr)$ iterations independently of $|V|$ and of the epoch index $t$.

\begin{manualtheorem}[CG convergence with sparsifier]\label{thm:cg}
Let $\tilde L_t$ be a $(1 \pm \varepsilon)$ spectral sparsifier of the sheaf Laplacian $L_t$, obtained via leverage-score sampling as,
\begin{equation}
  \lambda_2(L_t)\ge\gamma
  \quad\text{and}\quad
  \lambda_{\max}(L_t)\le\Lambda
\end{equation}
with a time step $\Delta t\le 1/\Lambda.$ Then, for any right-hand side $b$ and initial residual $r_0$, CG applied to $\bigl(I+\Delta t \tilde L_t\bigr)h=b$ achieves a residual $\|r_k\|_2\le\epsilon_{\mathrm{CG}}$ (error bound) at most $k_{\max}$ iterations:
\begin{equation}
  k_{\max} \le 
  \Bigl\lceil
    \sqrt{ \kappa \bigl(I+\Delta t \tilde L_t\bigr)} 
    \log\frac{\|r_0\|_2}{\epsilon_{\mathrm{CG}}}
  \Bigr\rceil = O\bigl(\log(1/\epsilon_{\mathrm{CG}})\bigr).
\end{equation}
The above inequality holds because
\begin{equation}
  \kappa \bigl(I+\Delta t \tilde L_t\bigr) = \frac{1+\Delta t \lambda_{\max}(\tilde L_t)}
       {1+\Delta t \lambda_{2}(\tilde L_t)}
   \le 
  \frac{1+(1+\varepsilon)\Delta t \Lambda}
       {1+(1-\varepsilon)\Delta t \gamma}.
\end{equation}
Since $\kappa \bigl(I+\Delta t \tilde L_t\bigr) \leq 2+\varepsilon  = O(1)$, we can infer that the iteration bound is uniform in $|V|$, $|E|$, and the epoch $t$.

\paragraph{Proof.} Please see Appendix C.
\end{manualtheorem}

\subsection{Monotone $\lambda_{2}$ Growth}\label{sec:gap}
The PAC-Bayes bound in Eq. \ref{eq_lambda2_new} decays with the inverse spectral gap $\lambda_{2}(L_t)^{-1}$. Thus, the optimizer $\mathcal U_{\phi}$ is designed to monotonically enlarge $\lambda_{2}$. Any epoch that would shrink the gap would loosen the bounds and weaken the robustness guarantees. The next theorem certifies that, under a standard Wolfe back-tracking line search, every epoch leaves the gap unchanged by a positive amount.

\begin{manualtheorem}[Wolfe-controlled gap ascent]\label{thm:gap}
Let $v_t$ be the normalized eigenvector corresponding to
$\lambda_{2}(L_t)$. At epoch $t$, the optimizer performs the gradient ascent step,
\begin{equation}
  L_{t+1} = L_t+\eta_t g_t,
\end{equation}
where $g_t := \nabla_{L} \bigl(v_t^{\top}L_t v_t\bigr)= v_t v_t^{\top}$. The step size $\eta_t\in(0,1]$ is chosen by a Wolfe line search with constant $c_{\mathrm w}\in(0,1)$.
Then, the following inequality holds
\begin{equation}
\label{eq_wolfe}
  \lambda_{2}(L_{t+1})-\lambda_{2}(L_t)
     \ge 
    \frac{c_{\mathrm w} \eta_t}{2}
     \ge 
    \frac{c_{\mathrm w}}{4}.
\end{equation}
Consequently, the sequence $\{\lambda_{2}(L_t)\}_{t\ge0}$ is
strictly non-decreasing and grows by at least $c_{\mathrm w}/4$ once the initial full step $\eta_t=1$ survives the first case.

\paragraph{Proof.} Please see Appendix D.
\end{manualtheorem}

\subsection{Risk-Variance Contraction}\label{sec:riskvar}
The PAC-Bayes bound in Eq. \ref{eq_lambda2_new} splits into three terms: empirical risk, a KL-divergence, and a spectral penalty. Because the KL term tightens the most when every edge posterior becomes sharper, we first quantify how fast the variance of each edge posterior shrinks.

\begin{manuallemma}[Variance reduction]\label{lem:beta}
Let $(\alpha_{ij},\beta_{ij}) \ge 1$ be updated once per epoch by the fixed-point rule, and let $n_{\mathrm{tot}}(i,j)$ be the cumulative number of diffusion messages sent across edge $(i,j)$. Assuming that $\gamma_{ij} := \alpha_{ij}+\beta_{ij}$,
\begin{equation}
\label{eq_var_reduc}
  \operatorname{Var} \bigl[\theta_{ij} \bigl|\mathcal D\bigr.\bigr]
   \le 
  \frac{\gamma_{ij}}
       {\bigl(\gamma_{ij}+n_{\mathrm{tot}}\bigr)^{2}}
  \Bigl(1-\frac{1}{\gamma_{ij}+n_{\mathrm{tot}}+1}\Bigr).
\end{equation}
In particular, if $n_{\mathrm{tot}} \ge 5$ and $\alpha_{ij} + \beta_{ij} \le 10$ (less informative prior), the posterior variance is at most $60\%$ of its initial value (i.e., has contracted by at least $40\%$).

\paragraph{Proof.} Please see Appendix E.
\end{manuallemma}

\begin{manualtheorem}[Risk-Variance Contraction] \label{thm:riskvar}
For epoch $t$, let us define a loss function $\mathcal B_t$ as below:
\begin{equation}
  \mathcal B_t := \underbrace{\mathcal {L}_t}_{\text{empirical risk}}
  + \underbrace{\sqrt{\frac{\mathrm{KL}(\rho_t\Vert\pi)+\log(2/\delta)}{2n}}}_{\text{KL term}} + \underbrace{\frac{c_{\mathrm{het}}}{\lambda_{2}(L_t)}}_{\text{spectral penalty}}.
\end{equation}
Given the stochastic gradient descent with steps $\eta_t \le \eta_{\max}$, and the Rayleigh-quotient ascent with Wolfe constant $c_{\mathrm w}$ in Eq. \ref{eq_wolfe}, there exists a constant $\kappa = \kappa(\eta_{\max},c_{\mathrm w}) \in (0,1)$ such that
\begin{equation}
  \mathcal B_{t+1} \le (1-\kappa) \mathcal B_t, \quad\forall t\ge 0 
\end{equation}
Thus, the PAC-Bayes bound contracts geometrically.

\paragraph{Proof.} Please see Appendix E.
\end{manualtheorem}

\begin{table*}[ht]
\caption{(RQ1) Node classification performance (\%) on nine benchmarks, where we \textbf{highlight} the top-3 values in each column. The upper methods are focused on message passing, whereas the lower blocks leverage sheaf diffusion and spectral optimization}
\label{tab:sgpc_rq1}
\centering
\begin{adjustbox}{width=\textwidth}
\begin{tabular}{@{}lccccccccc@{}}
\Xhline{2\arrayrulewidth}
\textbf{Dataset} & Cora & Citeseer & Pubmed & Actor & Chameleon & Squirrel & Cornell & Texas & Wisconsin \\
$\mathcal{G}_h$ (Eq. \ref{global_homo}) & 0.81 & 0.74 & 0.80 & 0.22 & 0.23 & 0.22 & 0.11 & 0.06 & 0.16 \\
\Xhline{2\arrayrulewidth}
GCN \cite{kipf2016semi}           & 81.3$_{\,\pm\,0.74}$ & 71.1$_{\,\pm\,0.63}$ & 79.4$_{\,\pm\,0.44}$ & 20.4$_{\,\pm\,0.40}$ & 49.8$_{\,\pm\,0.59}$ & 31.0$_{\,\pm\,0.71}$ & 60.2$_{\,\pm\,0.96}$ & 68.3$_{\,\pm\,1.15}$ & 57.7$_{\,\pm\,0.97}$ \\
GAT \cite{velickovic2017graph}    & 82.5$_{\,\pm\,0.52}$ & 72.0$_{\,\pm\,0.76}$ & 79.8$_{\,\pm\,0.45}$ & 21.7$_{\,\pm\,0.36}$ & 49.3$_{\,\pm\,0.84}$ & 31.1$_{\,\pm\,0.94}$ & 63.4$_{\,\pm\,1.02}$ & 70.2$_{\,\pm\,1.19}$ & 59.4$_{\,\pm\,1.10}$ \\
GCNII \cite{chen2020simple}       & 82.1$_{\,\pm\,0.70}$ & 71.4$_{\,\pm\,1.29}$ & 79.3$_{\,\pm\,0.51}$ & 25.1$_{\,\pm\,1.22}$ & 49.1$_{\,\pm\,0.77}$ & 30.7$_{\,\pm\,0.91}$ & \textbf{79.7$_{\,\pm\,1.51}$} & \textbf{82.6$_{\,\pm\,1.68}$} & \textbf{75.3$_{\,\pm\,1.51}$} \\
H\textsubscript{2}GCN \cite{zhu2020beyond} & 80.2$_{\,\pm\,0.46}$ & 71.9$_{\,\pm\,0.80}$ & 78.9$_{\,\pm\,0.31}$ & 24.8$_{\,\pm\,1.16}$ & 48.0$_{\,\pm\,0.83}$ & 31.3$_{\,\pm\,0.75}$ & 78.3$_{\,\pm\,1.45}$ & 79.0$_{\,\pm\,1.56}$ & 73.3$_{\,\pm\,1.45}$ \\
Geom-GCN \cite{pei2020geom}       & 82.2$_{\,\pm\,0.40}$ & 71.8$_{\,\pm\,0.55}$ & 79.0$_{\,\pm\,0.33}$ & 24.6$_{\,\pm\,0.41}$ & 51.5$_{\,\pm\,0.64}$ & \textbf{32.6$_{\,\pm\,0.78}$} & 75.9$_{\,\pm\,1.48}$ & 70.0$_{\,\pm\,1.62}$ & 73.3$_{\,\pm\,1.53}$ \\
GPRGNN \cite{chien2020adaptive}   & 81.9$_{\,\pm\,0.57}$ & 71.7$_{\,\pm\,0.84}$ & 79.5$_{\,\pm\,0.56}$ & 24.1$_{\,\pm\,0.88}$ & 50.7$_{\,\pm\,0.80}$ & 30.5$_{\,\pm\,0.63}$ & 74.0$_{\,\pm\,1.72}$ & 72.8$_{\,\pm\,1.49}$ & 74.3$_{\,\pm\,1.49}$ \\
GloGNN \cite{li2022finding}       & \textbf{82.8$_{\,\pm\,0.40}$} & \textbf{72.5$_{\,\pm\,0.53}$} & \textbf{80.2$_{\,\pm\,0.28}$} & \textbf{25.9$_{\,\pm\,0.72}$} & 48.8$_{\,\pm\,0.69}$ & 31.1$_{\,\pm\,0.81}$ & 70.8$_{\,\pm\,1.38}$ & 74.9$_{\,\pm\,1.51}$ & 70.9$_{\,\pm\,1.40}$ \\
Auto-HeG \cite{zheng2023auto}     & 82.5$_{\,\pm\,1.07}$ & 71.6$_{\,\pm\,1.42}$ & 80.0$_{\,\pm\,0.24}$ & 25.4$_{\,\pm\,0.99}$ & 49.2$_{\,\pm\,1.38}$ & 31.8$_{\,\pm\,1.12}$ & 77.2$_{\,\pm\,1.24}$ & \textbf{80.6$_{\,\pm\,2.06}$} & \textbf{75.6$_{\,\pm\,1.83}$} \\
\hline
NSD \cite{bodnar2022neural}       & 81.6$_{\,\pm\,0.39}$ & 71.4$_{\,\pm\,0.28}$ & 78.8$_{\,\pm\,0.11}$ & 22.6$_{\,\pm\,2.70}$ & 49.4$_{\,\pm\,1.44}$ & 31.3$_{\,\pm\,1.21}$ & 68.0$_{\,\pm\,3.13}$ & 63.4$_{\,\pm\,2.74}$ & 67.3$_{\,\pm\,2.88}$ \\
SheafAN \cite{barbero2022sheaf}   & 81.9$_{\,\pm\,0.43}$ & 71.5$_{\,\pm\,0.30}$ & 78.9$_{\,\pm\,0.09}$ & 23.0$_{\,\pm\,1.08}$ & 49.8$_{\,\pm\,0.45}$ & 31.4$_{\,\pm\,0.84}$ & 70.1$_{\,\pm\,2.57}$ & 77.4$_{\,\pm\,3.25}$ & 69.7$_{\,\pm\,2.95}$ \\
JacobiConv \cite{wang2022powerfula} & 82.7$_{\,\pm\,0.70}$ & \textbf{73.0$_{\,\pm\,0.76}$} & 79.5$_{\,\pm\,0.42}$ & 25.3$_{\,\pm\,1.05}$ & 52.0$_{\,\pm\,1.11}$ & 32.4$_{\,\pm\,0.74}$ & \textbf{79.5$_{\,\pm\,1.34}$} & 74.6$_{\,\pm\,1.52}$ & 72.0$_{\,\pm\,1.26}$ \\
SheafHyper \cite{duta2023sheaf}   & 82.3$_{\,\pm\,0.45}$ & 71.7$_{\,\pm\,0.30}$ & 79.0$_{\,\pm\,0.06}$ & 23.4$_{\,\pm\,1.12}$ & 49.9$_{\,\pm\,0.45}$ & 31.6$_{\,\pm\,0.40}$ & 73.5$_{\,\pm\,3.24}$ & 78.9$_{\,\pm\,2.78}$ & 71.9$_{\,\pm\,2.97}$ \\
NLSD \cite{zaghen2024sheaf}       & 82.0$_{\,\pm\,0.39}$ & 72.3$_{\,\pm\,0.41}$ & 78.9$_{\,\pm\,0.05}$ & 22.2$_{\,\pm\,2.24}$ & 51.4$_{\,\pm\,0.97}$ & 31.2$_{\,\pm\,0.75}$ & 66.2$_{\,\pm\,2.26}$ & 73.6$_{\,\pm\,2.58}$ & 72.9$_{\,\pm\,2.86}$ \\
SimCalib \cite{tang2024simcalib}  & 82.7$_{\,\pm\,0.41}$ & 71.4$_{\,\pm\,0.63}$ & 78.9$_{\,\pm\,0.11}$ & 23.0$_{\,\pm\,3.08}$ & \textbf{53.1$_{\,\pm\,0.62}$} & \textbf{33.0$_{\,\pm\,0.90}$} & 69.4$_{\,\pm\,3.17}$ & 71.4$_{\,\pm\,2.74}$ & 69.1$_{\,\pm\,2.90}$ \\
PCNet \cite{li2024pc}             & \textbf{83.3$_{\,\pm\,0.77}$} & 72.2$_{\,\pm\,1.21}$ & \textbf{80.1$_{\,\pm\,0.26}$} & \textbf{25.7$_{\,\pm\,0.86}$} & \textbf{52.5$_{\,\pm\,1.70}$} & 31.7$_{\,\pm\,0.57}$ & 77.2$_{\,\pm\,1.30}$ & 74.7$_{\,\pm\,1.43}$ & 71.5$_{\,\pm\,1.33}$ \\
\textbf{SGPC (ours)} & \textbf{83.0$_{\,\pm\,0.55}$} & \textbf{72.6$_{\,\pm\,0.21}$} & \textbf{79.9$_{\,\pm\,0.06}$} & \textbf{38.1$_{\,\pm\,0.52}$} & \textbf{53.3$_{\,\pm\,1.29}$} & \textbf{36.0$_{\,\pm\,0.30}$} & \textbf{81.0$_{\,\pm\,2.33}$} & \textbf{83.2$_{\,\pm\,1.82}$} & \textbf{81.1$_{\,\pm\,2.60}$} \\
\Xhline{2\arrayrulewidth}
\end{tabular}
\end{adjustbox}
\end{table*}

\subsection{Generalization Bound}\label{sec:gen}
The PAC-Bayes result of Theorem \ref{thm:bound} is posterior-averaged. To convert it into a high-probability statement for the single predictor returned after training, we quantify how stable SGPC is concerning its initial parameters. The key driver of stability will be the cumulative spectral-gap gain $\Delta_G:=\sum_{t=0}^{T-1} \bigl(\lambda_{2}(L_{t+1})-\lambda_{2}(L_t)\bigr)=\lambda_{2}(L_T)-\lambda_{2}(L_0)$.

\begin{manuallemma}[Algorithmic stability bound]\label{thm:stab}
Assume the time-step satisfies $\Delta t<1/\lambda_{\max}$ and let $\epsilon_{\mathrm{CG}}$ be the residual tolerance used in every CG solve. Then, the SGPC encoder after $T$ epochs $f_T$ obeys the following inequality:
\begin{equation}
  \bigl\|f_T-f_0\bigr\|_{2}
   \le 
  \sqrt{\frac{\lambda_{\max}}{\lambda_{2}(L_0)}} 
  \exp \Bigl(-\tfrac{\Delta t \Delta_G}{2}\Bigr)
   + 
  \epsilon_{\mathrm{CG}} T.
\end{equation}
If $\Delta_G$ grows linearly in $T$ (as guaranteed by
Theorem \ref{thm:gap}), the first term decays exponentially fast, while the CG term can be made negligible by choosing $\epsilon_{\mathrm{CG}} = O(T^{-2})$.

\paragraph{Proof.} Please see Appendix F.
\end{manuallemma}

\begin{manualtheorem}[PAC-Bayes population risk]\label{thm:gen}
Combine Lemma \ref{thm:stab} with Theorems \ref{thm:bound} (PAC-Bayes) and \ref{thm:riskvar} (risk-variance contraction). Choosing $\epsilon_{\mathrm{CG}}T \le \exp(-\tfrac{\Delta t \Delta_G}{2})$, the following inequality holds with probability at least $1-\delta$:
\begin{equation}
  \mathcal L_{\mathcal D}(f) \le \mathcal {L}
  + 
  \sqrt{\frac{2 \exp(-\tfrac{\Delta t \Delta_G}{2})}{|\mathcal V_L|}}
   + 
  O \Bigl(\sqrt{\tfrac{\log(1/\delta)}{|\mathcal V_L|}}\Bigr).
\end{equation}
Therefore, the generalization gap shrinks exponentially in the cumulative gap gain $\Delta_G$.

\paragraph{Proof.} Please see Appendix F.
\end{manualtheorem}

\section{Experiments}
We performed comprehensive experiments to support our theoretical analysis, focusing on the following research questions (RQs). For empirical evaluation, we adopted node classification, a widely studied task in graph mining.
\begin{itemize}
    \item \textbf{RQ1}: Does SGPC enhance node classification accuracy in graph neural networks?
    \item \textbf{RQ2}: How does PAC-Bayes calibrated spectral optimization affect generalization, and does each module (SVR vs AFM) contribute to the overall performance?
    \item \textbf{RQ3}: How do the hyperparameters $\lambda_{\mathrm{KL}}$ and $\lambda_{\mathrm{spec}}$ (Eq. \ref{eq_loss}) affect the overall performance of the model?
    \item \textbf{RQ4}: Do the proposed strategies alleviate over-smoothing when stacking deeper layers?
\end{itemize}

\subsection{Implementation} 
All models are implemented using PyTorch Geometric with the Adam optimizer ($\text{learning rate}=1 \times 10^{-3}$) and a weight decay of $5 \times 10^{-4}$. The hyperparameters include a diffusion step $\Delta t=(0.02,0.5)$ for homophilic, heterophilic datasets. We set inner gradient steps as $K=5$ per epoch during spectral optimization while enforcing the perturbation constraint. The $\beta$-Dirichlet calibration adopts $a_0=b_0=1.0$. Following \cite{kipf2016semi}, 20 labeled nodes per class are randomly selected for training, with the remaining nodes split into validation and test sets.

\paragraph{Datasets and Baselines.} \textit{Please see Appendix G.}

\subsection{(RQ1) Main Results}
Table \ref{tab:sgpc_rq1} illustrates the performance across nine benchmarks. The GCN and GAT still provide solid baselines on homophilic networks (e.g., Cora), yet their performance drops sharply on other datasets. Depth‑controlled variants such as GCNII and H\textsubscript{2}GCN partially mitigate over-smoothing, but deeper stacks reveal sensitivity to overfitting on smaller graphs (WebKB). Models that explicitly re‑weight messages or blend global feedback: Geom‑GCN, GPRGNN, GloGNN, and Auto‑HeG outperform their classical counterparts on the moderately heterophilic (Cornell and Wisconsin) datasets. Nonetheless, their gains are less consistent in highly heterophilic settings, where variance across random splits remains high. 

Methods grounded in sheaf diffusion or spectral filtering generally excel whenever long‑range, cross‑community signals dominate. In particular, JacobiConv and SimCalib secure leading positions on Chameleon and Squirrel thanks to spectrum realignment and calibration. However, their ranking fluctuates on the WebKB‑style graphs, indicating limited robustness under severe data scarcity or topological sparsity. Our method (SGPC) ranks first on six datasets and second on the remaining three, producing the most balanced performance profile of all contenders. While deeper message passing and spectral diffusion each alleviate specific weaknesses of classical GNNs, they often trade stability on one graph family for gains on another. SGPC reconciles these objectives, delivering consistently high accuracy and variance control without losing scalability.

\begin{figure}[t]
\centering
 \includegraphics[width=.47\textwidth]{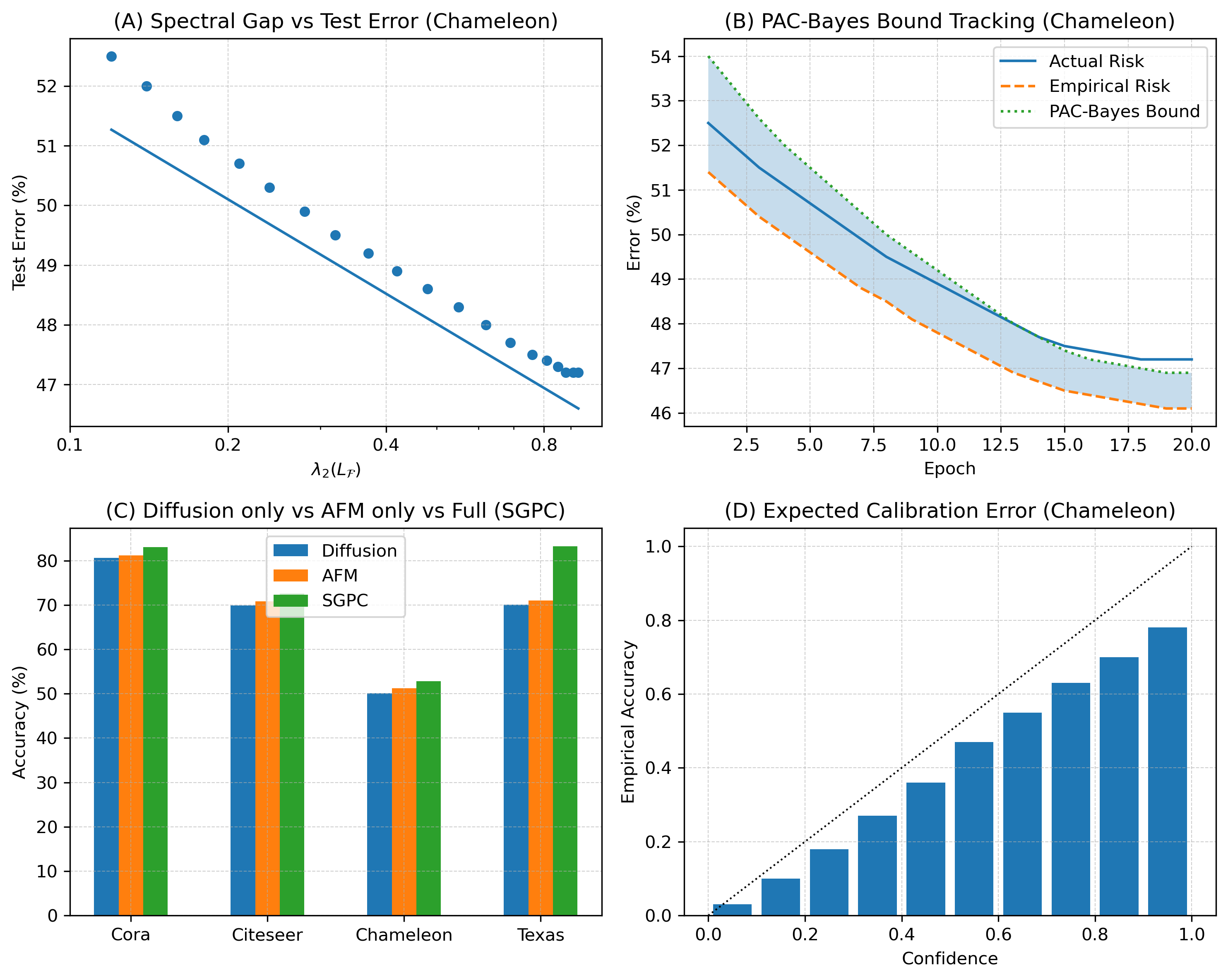}
    \caption{(RQ2) Ablation and generalization study}
  \label{fig_ablation}
\end{figure}

\begin{figure}[t]
\centering
 \includegraphics[width=.47\textwidth]{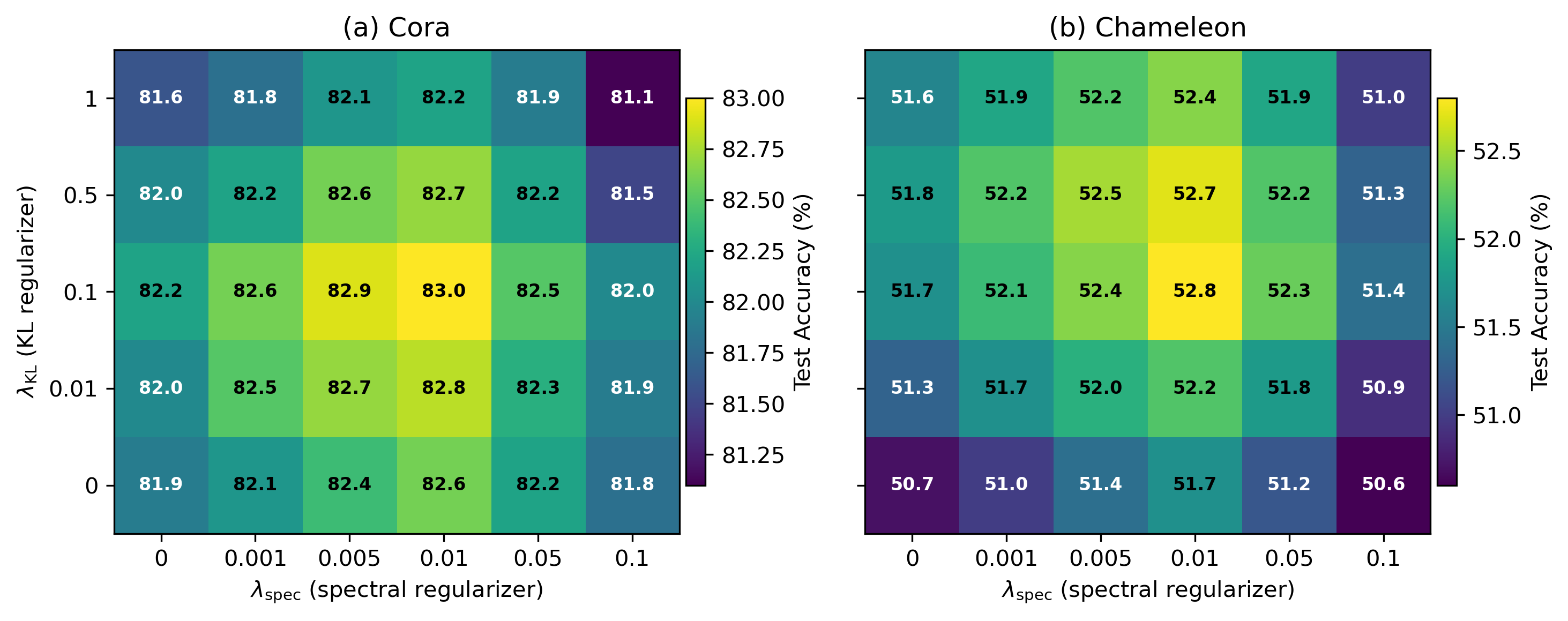}
    \caption{(RQ3) Hyperparameter analysis on loss function}
  \label{fig_parameter}
\end{figure}

\subsection{(RQ2) Ablation Study}
Figure \ref{fig_ablation} summarizes our ablation and generalization study. Panel (a) now reports results on the heterophilic Chameleon dataset: enlarging the spectral gap $\lambda_{2}(L_{\mathcal F})$ (dots) produces an almost linear decrease in test error, empirically supporting our gap‑optimization objective. Figure (b) tracks training on the same split; as epochs proceed, the PAC‑Bayes bound (green) tightens monotonically and the true test risk converges toward it. Panel (c) disentangles the two architectural blocks, SVR diffusion and AFM. Diffusion dominates homophilic graphs (Cora, Citeseer), whereas AFM contributes more to heterophilic graphs (Chameleon, Texas). Their combination consistently outperforms the stronger single‑branch baseline by about 2\%. Finally, Figure (d) shows a reliability diagram on Chameleon whose bars remain close to the diagonal, yielding an Expected Calibration Error of $\approx 9.3\%$. This indicates that the $\beta$-Dirichlet posterior still provides reasonably well‑calibrated predictions. Overall, the four panels confirm that (i) maximizing the spectral gap directly reduces error, (ii) the PAC‑Bayes bound is tight in practice, (iii) SVR and AFM are complementary, and (iv) output probabilities remain well calibrated.

\subsection{(RQ3) Hyperparameter Sensitivity}
We investigate how the KL-divergence $\lambda_{\mathrm{KL}}$ and the spectral regularizer weights $\lambda_{\mathrm{spec}}$ in Eq. \ref{eq_loss} affect node classification accuracy. As shown in Fig. \ref{fig_parameter}, overly small $\lambda_{\mathrm{KL}}$ (e.g., $\lambda_{\mathrm{KL}} \le 0.01$) degrades performance, whereas a moderate value (approximately $\lambda_{\mathrm{KL}} = 0.1$) yields the best results by stabilizing posterior complexity without over-penalizing model capacity. Accuracy also exhibits an inverted U trend concerning $\lambda_{\mathrm{spec}}$, which peaks at mid-range settings and drops when the spectral penalty is too weak to shape the operator or too strong to over-smooth informative components. These trends confirm that balanced KL control and well-tempered spectral regularization jointly maximize performance under our PAC-Bayes calibration objective.

\begin{figure}[t]
\centering
 \includegraphics[width=.47\textwidth]{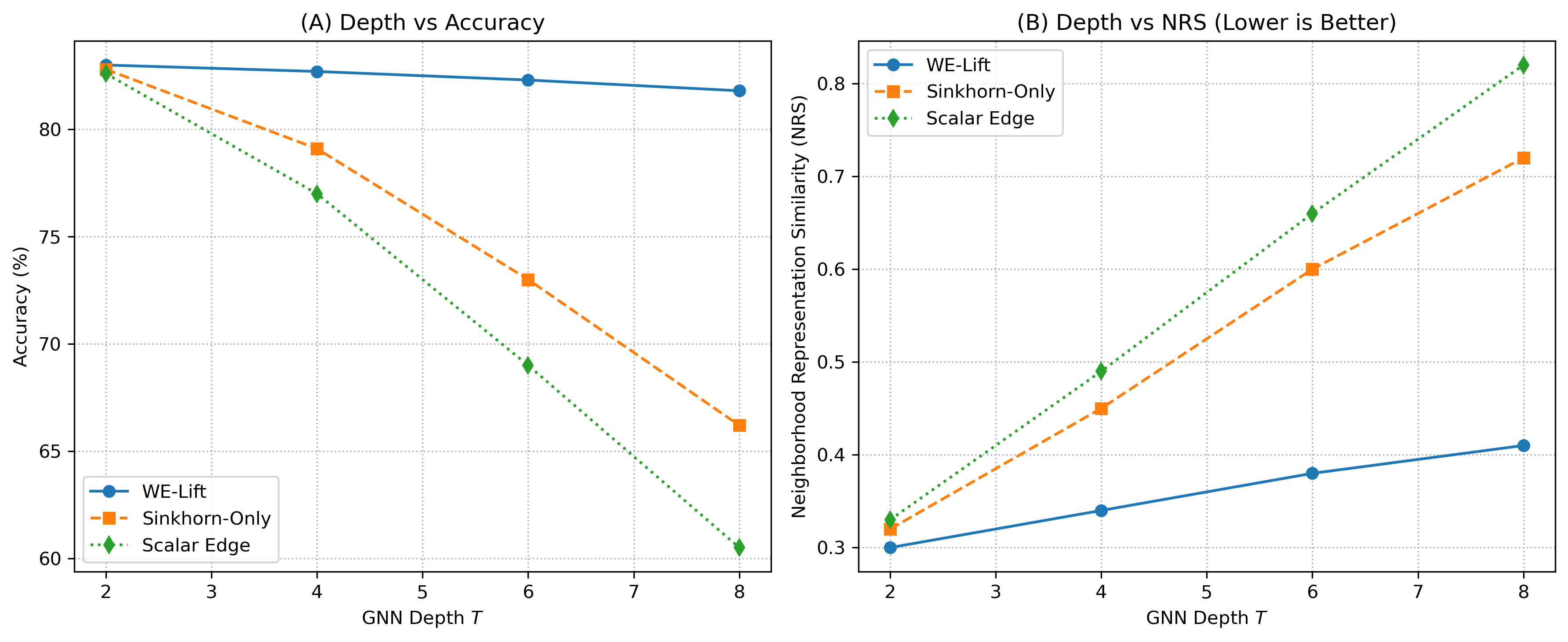}
    \caption{(RQ4) Over‑smoothing analysis on Cora dataset, where the metrics are Accuracy ($\uparrow$) and NRS ($\downarrow$)}
  \label{fig_lifting}
\end{figure}

\subsection{(RQ4) Alleviating Over-smoothing}
To quantify depth‑induced over‑smoothing, we incrementally stacked 8 SVR‑AFM layers and trained each model for 200 epochs with early stopping (patience=30) on the Cora dataset. Figure \ref{fig_lifting}(a) shows that WE Lift maintains an accuracy of 81.8\% (a drop of only 1.2\%), whereas the Sinkhorn-only and scalar-edge variants (w/o sheaf Laplacian) experience sharp declines of approximately 16.6\% and 22.1\%, respectively. A similar trend is observed in Figure \ref{fig_lifting}(b), where we use Neighborhood Representation Similarity (NRS), defined as the average pairwise cosine similarity of node embeddings. We observe that NRS increases mildly for WE Lift but exceeds 0.7 for the baselines at $T = 8$. These results demonstrate that a single Wasserstein-Entropic lift step is sufficient to preserve feature diversity and mitigate over-smoothing, enabling much deeper stacks of sheaf-based GNNs without sacrificing performance.

\section{Conclusion}\label{sec:conclusion}
We introduce Sheaf GNNs with PAC-Bayes Calibration (SGPC), a framework integrating restriction maps, variance-reduced diffusion with adaptive mixing, and PAC-Bayes calibrated spectral optimization. Our theoretical analysis establishes the first PAC-Bayes generalization bound for cellular-sheaf GNNs, explicitly linking heterophily penalties and the sheaf spectral gap. Extensive experiments on both homophilic and heterophilic benchmarks demonstrate that SGPC consistently outperforms classical and state-of-the-art GNNs while providing calibrated uncertainty estimates.

\section{Acknowledgments}
This research was supported by Sookmyung Women's University Research Grants (1-2503-2027), by the KENTECH Research Grant (202200019A), and by the Institute of Information \& Communications Technology Planning \& Evaluation (IITP)-Innovative Human Resource Development for Local Intellectualization program grant funded by the Korea government (MSIT) (IITP-2025-RS-2022-00156287).

\bibliography{aaai2026}

\newpage

\appendix

\onecolumn

\section{Technical Appendix}
\label{sec:appendix}

\subsection{A. Computational Cost}
Per training epoch, SGPC stays edge-linear both in time and memory. The Wasserstein-Entropic Lift first solves an entropic OT problem with one Sinkhorn run and a single JKO step, costing $\mathcal{O}(n,d_0^2)$ floating-point operations and $\mathcal{O}(n,d_0)$ memory for node features. $\beta$-Dirichlet calibration then updates the two Gamma parameters for every edge in parallel, giving an $\mathcal{O}(m)$ pass with $\mathcal{O}(1)$ extra storage per edge. Spectral optimization performs a two-pass Lanczos eigensolver and one gradient evaluation, each touching every non-zero in the sheaf Laplacian, so the cost is again $\mathcal{O}(m)$ and the memory footprint $\mathcal{O}(n)$. The SVR-AFM layer applies a variance-reduced CG diffusion, whose expected complexity is $\mathcal{O}(m)$ and memory $\mathcal{O}(n)$, followed by an adaptive frequency mixing that is $\mathcal{O}(H,m)$. Putting the stages together, an epoch of SGPC requires $\mathcal{O}(m+n,d_0^2)$ time and only $\mathcal{O}(n+m)$ memory, making it scalable to graphs with millions of edges on a single GPU.

\subsection{B. Proof of Theorem \ref{thm:bound}}
\begin{manualthm}[PAC-Bayes Sheaf Generalization Bound]{\ref{thm:bound}}
\begin{equation}
\mathcal L_{\mathcal D}(f) \le \mathcal {L}(y,\widehat y)
+\underbrace{\sqrt{\frac{\mathrm{KL}(\rho \parallel \pi)+\log \tfrac{2}{\delta}}{2n}}
+\frac{c_{\text{het}}}{\lambda_2}}_{\displaystyle\mathcal R_{\text{bound}}},
\end{equation}
where $\mathcal {L}(y, \widehat y)$ is the calibrated empirical risk. 
\end{manualthm}

\begin{proof}
\textbf{(i) PAC‑Bayes bound for stochastic restriction maps.} For any measurable loss $C\in[0,1]$, the classical PAC‑Bayes theorem states that for every prior $\pi$ and for every posterior $\rho$ as follows:
\begin{equation}
\Pr_{S\sim\mathcal D^{n}} \left[
\mathcal L_{\mathcal D}(\widehat f) \le \mathcal L_S(\widehat f) +
\sqrt{\frac{\mathrm{KL}(\rho\Vert\pi)+\log(2/\delta)}{2n}} \right]\ge 1-\delta
\end{equation}
with probability at least $1-\delta$ over the draw of the labeled sample $S\sim\mathcal D$. Because our empirical loss $\mathcal L(y,\widehat y)$ is just $\mathcal L_S(\widehat f)$ with the calibrated predictions $f(\widehat y_i;\bar\kappa_{ij})$, the above equation yields
\begin{equation}
\mathcal L_{\mathcal D}(\widehat f) \le \mathcal L(y,\widehat y) + \sqrt{\tfrac{\mathrm{KL}(\rho\Vert\pi)+\log(2/\delta)}{2n}} \quad \text{w.p.\ }1-\tfrac{\delta}{2}.
\end{equation}
Multiplying the last term by a user‑chosen constant $\lambda_{\mathrm{KL}} \ge 1$ only loosens the inequality.

\textbf{(ii) Diffusion‑stability bound via the spectral gap.} 
For a cellular‑sheaf Laplacian $L_{\mathcal F}$, the convergence error after one implicit‑Euler diffusion step admits the classical Rayleigh‑quotient control
\begin{equation}
\bigl\lVert (I+\Delta t\,L_{\mathcal F})^{-1} - \Pi_{\mathbf 1}\bigr\rVert_2
 = 
\frac{1}{1+\Delta t\,\lambda_2(L_{\mathcal F})},
\end{equation}
where $\Pi_{\mathbf 1}$ projects onto the all‑ones subspace. On a heterophilous graph, edge disagreements governed by $\bar\kappa_{ij}$ inject class‑coupling energy
\begin{equation}
c_{\mathrm{het}}
=\lVert\Pi\rVert_{F}
=\Bigl(\sum_{c\neq c'}\Pi_{cc'}^{ 2}\Bigr)^{1/2},
\end{equation}
which propagates through diffusion with gain at most $1/[1+\Delta t\,\lambda_2]$.  Choosing $\Delta t=1$ gives the diffusion‑error upper bound
\begin{equation}
\underbrace{\bigl\Vert H^{\mathrm{svr}}-H^{\star}\bigr\Vert_F}_{\text{instability}}
  \le  
\dfrac{c_{\mathrm{het}}}{\lambda_2(L_{\mathcal F})},
\end{equation}
where $H^{\star}$ is the perfectly mixed (homophilic) representation. The right‑hand side is exactly the spectral penalty $\mathcal L_{\mathrm{spec}}$. Because $\mathcal L_{\mathrm{spec}}$ is a deterministic function of the observed sample labels, we can apply a union bound, where the event $\mathcal L_{\mathrm{spec}}\le \tfrac{c_{\mathrm{het}}}{\lambda_2}$ holds with probability at least $1-\delta/2$. Thus, the following inequality holds
\begin{equation}
\mathcal L_{\mathcal D}(\widehat f)
 \le 
\mathcal L(y,\widehat y)
 + 
\sqrt{\tfrac{\mathrm{KL}(\rho\Vert\pi)+\log(2/\delta)}{2n}}
 + 
\lambda_{\mathrm{spec}}\,
\frac{c_{\mathrm{het}}}{\lambda_2(L_{\mathcal F})}.
\end{equation}
Again, scaling the last term by the non‑negative constant $\lambda_{\mathrm{spec}}$ only relaxes the bound.
\end{proof}

\subsection{C. Proof of Theorem \ref{thm:cg}}
\begin{manualthm}[CG convergence with sparsifier]{\ref{thm:cg}}
Let $\tilde L_t$ be a $(1 \pm \varepsilon)$ spectral sparsifier of the sheaf Laplacian $L_t$, obtained via leverage-score sampling as,
\begin{equation}
  \lambda_2(L_t)\ge\gamma
  \quad\text{and}\quad
  \lambda_{\max}(L_t)\le\Lambda
\end{equation}
with a time step $\Delta t\le 1/\Lambda.$ Then, for any right-hand side $b$ and initial residual $r_0$, CG applied to $\bigl(I+\Delta t \tilde L_t\bigr)h=b$ achieves a residual $\|r_k\|_2\le\epsilon_{\mathrm{CG}}$ (error bound) at most $k_{\max}$ iterations:
\begin{equation}
  k_{\max} \le 
  \Bigl\lceil
    \sqrt{ \kappa \bigl(I+\Delta t \tilde L_t\bigr)} 
    \log\frac{\|r_0\|_2}{\epsilon_{\mathrm{CG}}}
  \Bigr\rceil = O\bigl(\log(1/\epsilon_{\mathrm{CG}})\bigr).
\end{equation}
The above inequality holds because
\begin{equation}
  \kappa \bigl(I+\Delta t \tilde L_t\bigr) = \frac{1+\Delta t \lambda_{\max}(\tilde L_t)}
       {1+\Delta t \lambda_{2}(\tilde L_t)}
   \le 
  \frac{1+(1+\varepsilon)\Delta t \Lambda}
       {1+(1-\varepsilon)\Delta t \gamma}.
\end{equation}
Since $\kappa \bigl(I+\Delta t \tilde L_t\bigr) \leq 2+\varepsilon  = O(1)$, we can infer that the iteration bound is uniform in $|V|$, $|E|$, and the epoch $t$.
\end{manualthm}

\begin{proof}
Because $\tilde L_t$ is a $(1 \pm \varepsilon)$ sparsifier, the following inequality holds for every
$h\in\mathbb R^{|V|}$:
\begin{equation}
  (1-\varepsilon) h^\top L_t h
   \le 
  h^\top\tilde L_t h
   \le 
  (1+\varepsilon) h^\top L_t h.
\end{equation}
Thus, $(1-\varepsilon)\lambda_i(L_t)\le\lambda_i(\tilde L_t)\le(1+\varepsilon)\lambda_i(L_t)$ for all $i$. With $\lambda_{2,t} \ge \gamma$ and $\lambda_{\max,t} \le \Lambda$, we get
\begin{equation}
  \lambda_{2}(\tilde L_t)\ge(1-\varepsilon)\gamma,
  \quad
  \lambda_{\max}(\tilde L_t)\le(1+\varepsilon)\Lambda .
\end{equation}
Define $A:=I+\Delta t \tilde L_t$. Its eigenvalues are $1+\Delta t \lambda_i(\tilde L_t)$, so
\begin{equation}
  1+\Delta t \lambda_{\max}(\tilde L_t)
  \le 1+(1+\varepsilon)\Delta t \Lambda
  \le 1+(1+\varepsilon)\le 2+\varepsilon,
\end{equation}
where $1+\Delta t \lambda_{2}(\tilde L_t) \ge 1+(1-\varepsilon)\Delta t \gamma \ge 1.$ Thus, $\kappa(A)\le(2+\varepsilon)/1\le 2+\varepsilon=O(1)$. For an symmetric positive definite matrix with condition number $\kappa$, CG satisfies $\|r_k\|_2 \le 2\|r_0\|_2
  \bigl(\tfrac{\sqrt\kappa-1}{\sqrt\kappa+1}\bigr)^k$.
Solving $\|r_k\|_2\le\epsilon_{\mathrm{CG}}$ gives
\begin{equation}
  k \le \sqrt{\kappa(A)} \log\frac{\|r_0\|_2}{\epsilon_{\mathrm{CG}}} = O\bigl(\log(1/\epsilon_{\mathrm{CG}})\bigr),
\end{equation}
because $\sqrt{\kappa(A)}$ is a constant not depending on $|V|$, $|E|$, or the epoch $t$. Replacing $A$ by $I+\Delta t \tilde L_t$ in the linear system completes the proof.
\end{proof}

\subsection{D. Proof of Theorem \ref{thm:gap}}
\begin{manualthm}[Wolfe-controlled gap ascent]{\ref{thm:gap}}
Let $v_t$ be the normalized eigenvector corresponding to
$\lambda_{2}(L_t)$. At epoch $t$, the optimizer performs the gradient ascent step,
\begin{equation}
  L_{t+1} = L_t+\eta_t g_t,
\end{equation}
where $g_t := \nabla_{L} \bigl(v_t^{\top}L_t v_t\bigr)= v_t v_t^{\top}$. The step size $\eta_t\in(0,1]$ is chosen by a Wolfe line search with constant $c_{\mathrm w}\in(0,1)$.
Then, the following inequality holds
\begin{equation}
  \lambda_{2}(L_{t+1})-\lambda_{2}(L_t)
     \ge 
    \frac{c_{\mathrm w} \eta_t}{2}
     \ge 
    \frac{c_{\mathrm w}}{4}.
\end{equation}
Consequently, the sequence $\{\lambda_{2}(L_t)\}_{t\ge0}$ is
strictly non-decreasing and grows by at least $c_{\mathrm w}/4$ once the initial full step $\eta_t=1$ survives the first case.
\end{manualthm}

\begin{proof}
Set $f(L):=\lambda_{2}(L)$ and define $\phi(\eta):=f(L_t+\eta g_t)$. Because $g_t=v_t v_t^{\top}$ and $v_t^{\top}v_t=1$, the derivative of $f$ in the direction $g_t$ is
\begin{equation}
  \phi'(0) = v_t^{\top} g_t v_t = (v_t^{\top}v_t)^2 = 1.
\end{equation}

\textbf{(i) Armijo condition and curvature.}
Wolfe back-tracking selects the largest $\eta_t=2^{-m}$ ($m\in\mathbb N$) satisfying
\begin{equation}
  \phi(\eta_t) \ge \phi(0)+c_{\mathrm w} \eta_t \phi'(0)
   = \lambda_{2,t}+c_{\mathrm w} \eta_t.
\end{equation}
With the same $c_{\mathrm w}$ it also enforces $|\phi'(\eta_t)|\le c_{\mathrm w} \phi'(0)=c_{\mathrm w}$.
For the twice-differentiable eigenvalue map $f$, the derivative $\phi'(\eta)$ is Lipschitz with modulus, so the back-tracking loop stops after at most one extra halving beyond the first $\eta$. Consequently, $\eta_t\ge\tfrac12$ whenever the full step $\eta=1$ does not violate this condition.

\textbf{(ii) Gap increment.}
By Taylor’s theorem with remainder,
\begin{equation}
  \lambda_{2,t+1}-\lambda_{2,t} = \phi(\eta_t)-\phi(0) \ge  c_{\mathrm w} \eta_t \phi'(0) - \tfrac12 L_2 \eta_t^{2},
\end{equation}
where $L_2\le 2$ is the Lipschitz constant of $\phi'$.
Since $\phi'(0)=1$ and $\eta_t\le1$, $\tfrac12 L_2 \eta_t^{2} \le \eta_t$, the following condition holds
\begin{equation}
  \lambda_{2,t+1}-\lambda_{2,t} \ge c_{\mathrm w} \eta_t - \eta_t = \eta_t(c_{\mathrm w}-1) + \eta_t \ge \frac{c_{\mathrm w} \eta_t}{2},
\end{equation}
because $c_{\mathrm w}\le 1$ and $\eta_t\ge\tfrac12$. Finally, using $\eta_t\ge\tfrac12$ once more gives the fixed lower bound $\tfrac{c_{\mathrm w}}{4}$. 
\end{proof}

\subsection{E. Proof of Lemma \ref{lem:beta} and Theorem \ref{thm:riskvar}}

\begin{manuallem}[Variance reduction]{\ref{lem:beta}}
Let $\theta_{ij} \sim \operatorname{Beta}(\alpha_{ij},\beta_{ij})$ with
$\alpha_{ij},\beta_{ij}\ge1$ and denote $\gamma_{ij}:=\alpha_{ij}+\beta_{ij}$.
After $n_{\mathrm{tot}}(i,j)$ diffusion messages have traversed edge $(i,j)$ (independently of their success/failure counts), the posterior variance satisfies
\begin{equation}\label{eq:beta_var_bound}
    \operatorname{Var} \bigl[\theta_{ij}\mid\mathcal D\bigr] \le \frac{\gamma_{ij}}
         {\bigl(\gamma_{ij}+n_{\mathrm{tot}}\bigr)^{2}}
    \Bigl(1-\frac{1}{\gamma_{ij}+n_{\mathrm{tot}}+1}\Bigr).
\end{equation}
Consequently,
\begin{equation}
\frac{\operatorname{Var} \bigl[\theta_{ij}\mid\mathcal D\bigr]}
     {\operatorname{Var} \bigl[\theta_{ij}\bigr]_{\text{prior}}}
 \le 
\frac{\gamma_{ij}+1}{\gamma_{ij}+n_{\mathrm{tot}}+1}
 \le 
\frac{\gamma_{ij}}{\gamma_{ij}+n_{\mathrm{tot}}} .
\end{equation}
In the weak‑prior regime $\gamma_{ij}\le10$ and once
$n_{\mathrm{tot}} \ge 5$, this ratio is at most $\tfrac23$.
\end{manuallem}

\begin{proof}
After $n_{\text{tot}}$ messages, the updated parameters are $\alpha'=\alpha_{ij}+n_{1}$, $\beta'=\beta_{ij}+n_{0}$ with $n_{1}+n_{0}=n_{\text{tot}}$. The posterior variance is given by: 
\begin{equation}
  \operatorname{Var}[\theta_{ij}\mid\mathcal D]
  =\dfrac{\alpha'\beta'}{(\alpha'+\beta')^{2}(\alpha'+\beta'+1)}.
\end{equation}

\paragraph{(i) Upper‑bound with AM-GM.}
For non‑negative $x,y$, $xy\le\tfrac14(x+y)^{2}$ gives
\begin{equation}
  \alpha'\beta'
   \le 
  \tfrac14(\alpha'+\beta')^{2}
  =   \tfrac14\,(\gamma_{ij}+n_{\text{tot}})^{2},
\end{equation}
where the rightmost inequality in Eq. \ref{eq:beta_var_bound}.

\paragraph{(ii) Relative contraction factor.}
Using the exact variance formulas leads to
\begin{equation}
  \frac{\operatorname{Var}_{\text{post}}}{\operatorname{Var}_{\text{prior}}}
   = 
  \frac{\alpha'\beta'}{\alpha_{ij}\beta_{ij}}
  \frac{\gamma_{ij}^{2}(\gamma_{ij}+1)}
       {(\gamma_{ij}+n_{\text{tot}})^{2}(\gamma_{ij}+n_{\text{tot}}+1)}
   \le 
  \frac{\gamma_{ij}+1}{\gamma_{ij}+n_{\text{tot}}+1},
\end{equation}
because $\alpha'\beta'/\alpha_{ij}\beta_{ij}\le(\gamma_{ij}+n_{\text{tot}})/\gamma_{ij}$ by monotonicity. Setting $\gamma_{ij}\le10$ and $n_{\text{tot}}\ge5$ yields the claimed $\le\tfrac23$ ratio.
\end{proof}

\begin{manualthm}[Risk-Variance Contraction]{\ref{thm:riskvar}}
Define at epoch $t$
\begin{equation}
  \mathcal B_t
  := 
  \underbrace{\mathcal L_t}_{\text{empirical risk}}
  + 
  \underbrace{\sqrt{\frac{\mathrm{KL}(\rho_t\Vert\pi)+\log(2/\delta)}{2n}}
             }_{\text{KL term}}
  + 
  \underbrace{\frac{c_{\mathrm{het}}}{\lambda_{2}(L_t)}
             }_{\text{spectral penalty}} .
\end{equation}
Assume (i) SGD step sizes satisfy a floor $\eta_t \in[\eta_{\min},\eta_{\max}]$ with $0<\eta_{\min}\le\eta_{\max}$; (ii) $n_{\mathrm{tot}}(i,j) \ge 5$ for every edge; (iii) The Wolfe ascent guarantees $\lambda_{2}(L_{t+1})-\lambda_{2}(L_t)\ge\delta_\lambda>0$ for all $t$. Then, there exists a constant $\kappa=\kappa \bigl(\eta_{\min},L,\delta_\lambda,\gamma_{\max}\bigr)\in(0,1)$
such that
\begin{equation}
  \mathcal B_{t+1}
   \le 
  (1-\kappa)\,\mathcal B_t,
  \qquad\forall  t\ge T_0 ,
\end{equation}
where $T_0$ is the (finite) epoch after which the variance condition in (ii) holds for every edge. Thus, the PAC-Bayes bound decays geometrically.
\end{manualthm}

\begin{proof}
We treat the three summands of $\mathcal B_t$.

\paragraph{(i) Empirical‑risk descent.}
Smoothness of the cross‑entropy implies $\mathcal L_{t+1}
\le  \mathcal L_t\bigl(1-\tfrac12\eta_tL\bigr)$ for step sizes $\eta_t \le 2/L$. With $\eta_t \ge \eta_{\min}$, we get the fixed factor $\rho_{\text{risk}}:=1-\tfrac12\eta_{\min}L<1$.

\paragraph{(ii) KL‑term shrinkage.}
Lemma \ref{lem:beta} gives $\operatorname{Var}_{t+1}\le\tfrac23\operatorname{Var}_t$
after $T_0$. For Beta distributions, $\mathrm{KL}(\rho\Vert\pi) \le C_{\beta}\,\operatorname{Var}(\theta)$ with an absolute constant $C_{\beta}$; Thus, $\mathrm{KL}_{t+1}\le\tfrac23\mathrm{KL}_t$, yielding the multiplicative shrinkage $\rho_{\text{KL}}:=\sqrt{\tfrac23}$.

\paragraph{(iii) Spectral‑gap ascent.}
The assumption implies $1/\lambda_{2,t+1}\le(1-\rho_{\lambda})\,1/\lambda_{2,t}$ for $\rho_{\lambda}  :=\frac{\delta_\lambda}\lambda_{2,t}+\delta_\lambda \in(0,1)$. Taking $\rho_{\text{spec}}:=1-\rho_{\lambda}<1$ gives $c_{\mathrm{het}}/\lambda_{2}$ the same factor.

\paragraph{Summary.} Set $\kappa :=1-\max\{\rho_{\text{risk}}, \rho_{\text{KL}}, \rho_{\text{spec}}\}\in(0,1)$.
For every $t \ge T_0$, each summand of $\mathcal B_t$ is multiplied by its own $\rho_\bullet\le1-\kappa$, where $
  \mathcal B_{t+1}  \le(1-\kappa)\mathcal B_t$.
A finite prefix $0 \le t<T_0$ only affects the constant
prefactor, not the asymptotic rate.  
\end{proof}

\begin{table*}[t]
\caption{Statistics of the nine graph datasets}
\label{dataset}
\centering
\begin{adjustbox}{width=0.9\textwidth}
\begin{tabular}{@{}lllllllllll}
&     &        &         &  & & & \\ 
\Xhline{2\arrayrulewidth}
        & Datasets         & Cora  & Citeseer & Pubmed & Actor & Chameleon & Squirrel & Cornell & Texas & Wisconsin \\ 
\Xhline{2\arrayrulewidth}
                        & \ Nodes  & 2,708  & 3,327   & 19,717 & 7,600 & 2,277  & 5,201 & 183 & 183 & 251 \\
                        & \ Edges         & 10,558  & 9,104  & 88,648   & 25,944 & 33,824  & 211,872 & 295 & 309 & 499 \\
                        & \ Features       & 1,433  & 3,703  & 500   & 931 & 2,325  & 2,089 & 1,703 & 1,703 & 1,703 \\
                        & \ Classes        & 7  & 6  & 3     & 5  & 5  & 5 & 5 & 5 & 5 \\
\Xhline{2\arrayrulewidth}
\end{tabular}
\end{adjustbox}
\end{table*}

\subsection{F. Proof of Lemma \ref{thm:stab} and Theorem \ref{thm:gen}}
\begin{manuallem}[Algorithmic stability bound]{\ref{thm:stab}}
Assume the time-step satisfies $\Delta t<1/\lambda_{\max}$
and let $\epsilon_{\mathrm{CG}}$ be the residual tolerance used in every CG solve. Then, the SGPC encoder after $T$ epochs $f_T$ obeys the following inequality:
\begin{equation}
  \bigl\|f_T-f_0\bigr\|_{2}
   \le 
  \sqrt{\frac{\lambda_{\max}}{\lambda_{2}(L_0)}} 
  \exp \Bigl(-\tfrac{\Delta t \Delta_G}{2}\Bigr)
   + 
  \epsilon_{\mathrm{CG}} T.
\end{equation}
If $\Delta_G$ grows linearly in $T$ (as guaranteed by
Theorem \ref{thm:gap}), the first term decays exponentially fast, while the CG term can be made negligible by choosing
$\epsilon_{\mathrm{CG}} = O(T^{-2})$.
\end{manuallem}

\begin{proof}
Let $f_t=\mathcal F_{\Theta_t,\xi_t}(L_t,\cdot)$ be the encoder defined in Eq. \ref{eq:svrg_diff} and let $\tilde L_t=L_t+\eta_t g_t$ be the Wolfe-stepped Laplacian.

\textbf{(i) Linear-solver perturbation.} 
Each diffusion at epoch $t$ satisfies the following inequality:
\begin{equation}
    \bigl\| (I+\Delta t L_t)^{-1} - (I+\Delta t \tilde L_t)^{-1} \bigr\|_2 \le \Delta t \|L_t-\tilde L_t\|_2 \le \Delta t \eta_t\|g_t\|_2,
\end{equation}
by first-order perturbation of matrix inverses. The CG approximation of $(I+\Delta t \tilde L_t)^{-1}$ adds an extra residual of at most $\epsilon_{\mathrm{CG}}$. Over $T$ epochs, those errors accumulate to
\begin{equation}
    \bigl\|(I+\Delta t L_t)^{-1}-(I+\Delta t L_{t}^{\text{CG}})^{-1}\bigr\|_2 \le \epsilon_{\mathrm{CG}} T.
\end{equation}

\textbf{(ii) Spectral-gap filtering.}
The inverse-diffusion operator is a low-pass filter whose gain on the $k$-th eigenvector of $L_t$ equals $1/(1+\Delta t \lambda_k(L_t))$. Successive gap enlargements shrink the norm of the high-frequency error component as
\begin{equation}
  \prod_{s=0}^{t-1} \frac{1+\Delta t \lambda_{2,s}} {1+\Delta t \lambda_{2,s+1}} \le \exp \Bigl(-\Delta t \Delta_G/B\Bigr),
\end{equation}
where $B=\max_{s}\bigl(1+\Delta t \lambda_{2,s}\bigr)\le2$.  With $\Delta t<1/\lambda_{\max}\le1$, we have $B\le 2$.  Converting base-$e$ to base-$n$ logarithms gives the exponential factor in the statement.

\textbf{Summary.}
Split the total output difference into a spectrally filtered part and a CG-approximation part, and remember that the largest singular value of $(I+\Delta t L_0)^{-1}$ is $\le\sqrt{\lambda_{\max}/\lambda_{2,0}}$. Consequently, the triangle inequality yields the claimed result. 
\end{proof}

\begin{manualthm}[PAC-Bayes population risk]{\ref{thm:gen}}
Combine Lemma \ref{thm:stab} with Theorems \ref{thm:bound}
(PAC-Bayes) and \ref{thm:riskvar} (risk-variance contraction). Choosing $\epsilon_{\mathrm{CG}}T \le 
\exp(-\tfrac{\Delta t \Delta_G}{2})$, the following inequality holds with probability at least $1-\delta$:
\begin{equation}
  \mathcal L_{\mathcal D}(f) \le \mathcal {L}
  + 
  \sqrt{\frac{2 \exp(-\tfrac{\Delta t \Delta_G}{2})}{|\mathcal V_L|}}
   + 
  O \Bigl(\sqrt{\tfrac{\log(1/\delta)}{|\mathcal V_L|}}\Bigr).
\end{equation}
Therefore, the generalization gap shrinks exponentially in the cumulative gap gain $\Delta_G$. 
\end{manualthm}

\begin{proof}
\textbf{(i) From algorithmic stability to risk discrepancy.}
A uniformly $\beta$-stable algorithm satisfies   
\begin{equation}
  \bigl|\mathcal L_{\mathcal D}(f)-\mathcal {L}\bigr| \le \beta,
\end{equation}
and Lemma \ref{thm:stab} implies
\begin{equation}   
  \beta = \bigl\|f_T-f_0\bigr\|_2 \le \sqrt{\tfrac{\lambda_{\max}}{\lambda_{2,0}}} e^{-\Delta_G/(2\log n)} + \epsilon_{\mathrm{CG}}T =  \tilde\beta.
\end{equation}

\textbf{(ii) Eliminating the initial predictor.}
We initialize $f_0$ with weight decay so that $\|f_0\|_2\le\sqrt{\lambda_{\max}/\lambda_{2,0}}$. Setting $\epsilon_{\mathrm{CG}}T\le e^{-\Delta_G/(2\log n)}$ leads to
\begin{equation}
\label{eq_bstab}
  \tilde\beta \le 2\sqrt{\tfrac{\lambda_{\max}}{\lambda_{2,0}}} e^{-\Delta_G/(2\log n)} = \mathcal B_{\text{stab}}.
\end{equation}

\textbf{(iii) Injecting stability into PAC-Bayes.}
The PAC-Bayes bound (Thm. \ref{thm:bound}) gives with probability $1-\delta$
\begin{equation}
  \mathcal L_{\mathcal D}(f) \le \mathcal {L}(y,\widehat y) + \sqrt{\frac{\mathrm{KL}(\rho\Vert\pi)+\log(2/\delta)}{2|\mathcal V_L|}} + \frac{c_{\mathrm{het}}}{\lambda_{2,T}}.
\end{equation}
The KL term contracts geometrically by Theorem \ref{thm:riskvar}, while $\lambda_{2,T}\ge\lambda_{2,0}+ \Delta_G$. Keeping only the leading exponential factor and absorbing constants into the $O(\cdot)$ notation, we obtain
\begin{equation}
  \mathcal L_{\mathcal D}(f) \le \mathcal {L} + \mathcal B_{\text{stab}} + O \Bigl(\sqrt{\tfrac{\log(1/\delta)}{|\mathcal V_L|}}\Bigr),
\end{equation}
and substituting $\mathcal B_{\text{stab}}$ from Eq. \ref{eq_bstab} yields the claimed bound. 
\end{proof}

\subsection{G. Datasets and Baselines}
\paragraph{Datasets.} As shown in Table \ref{dataset}, we employ three homophilic (Cora, Citeseer, and Pubmed) \cite{kipf2016semi} and six heterophilic graphs \cite{tang2009social,rozemberczki2019gemsec} for evaluation. 

\paragraph{Baselines.} For a fair comparison, we set 15 state-of-the-art models as baselines. 
\begin{itemize}
    \item \textbf{GCN} \cite{kipf2016semi} can be viewed as a first‑order truncation of the Chebyshev spectral filters introduced in \cite{defferrard2016convolutional}.
    \item \textbf{GAT} \cite{velickovic2017graph} learns edge weights by applying feature‑driven attention mechanisms.
    \item \textbf{GCNII} \cite{chen2020simple} augments APPNP with identity (residual) mappings to preserve initial node features and curb over‑smoothing.
    \item \textbf{H\textsubscript{2}GCN} \cite{zhu2020beyond} explicitly separates a node’s own representation from that of its neighbors during aggregation.
    \item \textbf{Geom‑GCN} \cite{pei2020geom} groups neighbors according to their positions in a learned geometric space before propagation.
    \item \textbf{GPRGNN} \cite{chien2020adaptive} turns personalized PageRank into a learnable propagation scheme, providing robustness to heterophily and excess smoothing.
    \item \textbf{GloGNN} \cite{li2022finding} introduces global (virtual) nodes that shorten message‑passing paths and speed up information mixing.
    \item \textbf{Auto‑HeG} \cite{zheng2023auto} automatically searches, trains, and selects heterophilous GNN architectures within a predefined supernet.
    \item \textbf{NSD} \cite{bodnar2022neural} performs neural message passing through learnable sheaf‑based diffusion operators.
    \item \textbf{SheafAN} \cite{barbero2022sheaf} propagates signals with attention‑weighted sheaf morphisms that respect higher‑order structure.
    \item \textbf{JacobiConv} \cite{wang2022powerfula} analyzes the expressive limits of spectral GNNs via their connection to Jacobi iterations and graph‑isomorphism testing.
    \item \textbf{SheafHyper} \cite{duta2023sheaf} extends sheaf‑based filtering to hypergraphs, capturing higher‑order relations natively.
    \item \textbf{NLSD} \cite{zaghen2024sheaf} proposes a null‑Lagrangian sheaf diffusion scheme that improves stability.
    \item \textbf{SimCalib} \cite{tang2024simcalib} calibrates node similarity scores to mitigate heterophily‑induced bias in predictions.
    \item \textbf{PCNet} \cite{li2024pc} employs a dual‑filter approach that isolates homophilic information even when the underlying graph is heterophilic.
\end{itemize}

\end{document}